\documentclass[sigconf,natbib=true,anonymous=false,screen,]{acmart}

\AtBeginDocument{%
  }

\usepackage{multirow}
\usepackage{booktabs}
\usepackage{makecell}

\usepackage{xcolor,colortbl}
\usepackage{acronym}
\usepackage{bm}
\usepackage[skip=1mm]{caption}
\usepackage{natbib}
\usepackage{hyperref}

\usepackage{enumitem}
\hypersetup{
  colorlinks   = true,    
  urlcolor     = blue,    
  linkcolor    = blue,    
  citecolor    = red      
}

\copyrightyear{2025}
\acmYear{2025}
\setcopyright{cc}
\setcctype{by}
\acmConference[RecSys '25]{Proceedings of the Nineteenth ACM Conference on Recommender Systems}{September 22--26, 2025}{Prague, Czech Republic}
\acmBooktitle{Proceedings of the Nineteenth ACM Conference on Recommender Systems (RecSys '25), September 22--26, 2025, Prague, Czech Republic}\acmDOI{10.1145/3705328.3748155}
\acmISBN{979-8-4007-1364-4/2025/09}

\begin{document}


\title[Rethinking the Privacy of Text Embeddings: A Reproducibility Study of Vec2Text]{Rethinking the Privacy of Text Embeddings: A Reproducibility Study of ``Text Embeddings Reveal (Almost) As Much As Text''}

\author{Dominykas Seputis}
\orcid{0009-0004-9720-1788}
\authornote{Both authors contributed equally to this research.}
\affiliation{%
  \institution{University of Amsterdam}
  \city{Amsterdam}
  \country{Netherlands}
}
\email{dominykas.seputis@student.uva.nl}

\author{Yongkang Li}
\orcid{0000-0001-6837-6184}
\authornotemark[1]
\authornote{Corresponding author.}
\affiliation{%
  \institution{University of Amsterdam}
  \city{Amsterdam}
  \country{Netherlands}
}
\email{y.li7@uva.nl}

\author{Karsten Langerak}
\orcid{0009-0008-6673-0846}
\affiliation{%
  \institution{University of Amsterdam}
  \city{Amsterdam}
  \country{Netherlands}
}
\email{karsten.langerak@student.uva.nl}

\author{Serghei Mihailov}
\orcid{0009-0005-4245-497X}
\affiliation{%
  \institution{University of Amsterdam}
  \city{Amsterdam}
  \country{Netherlands}
}
\email{serghei.mihailov@student.uva.nl}


\begin{abstract}

Text embeddings are fundamental to many natural language processing~(NLP) tasks, extensively applied in domains such as recommendation systems and information retrieval~(IR). Traditionally, transmitting embeddings instead of raw text has been seen as privacy-preserving. 
However, recent methods such as Vec2Text challenge this assumption by demonstrating that controlled decoding can successfully reconstruct original texts from black-box embeddings. The unexpectedly strong results reported by Vec2Text motivated us to conduct further verification, particularly considering the typically non-intuitive and opaque structure of high-dimensional embedding spaces. 
In this work, we reproduce the Vec2Text framework and evaluate it from two perspectives: (1) validating the original claims, and (2) extending the study through targeted experiments. First, we successfully replicate the original key results in both in-domain and out-of-domain settings, with only minor discrepancies arising due to missing artifacts, such as model checkpoints and dataset splits.
Furthermore, we extend the study by conducting a parameter sensitivity analysis, evaluating the feasibility of reconstructing sensitive inputs (e.g., passwords), and exploring embedding quantization as a lightweight privacy defense.
Our results show that Vec2Text is effective under ideal conditions, 
capable of reconstructing even password-like sequences that lack clear semantics.
However, we identify key limitations, including its sensitivity to input sequence length. We also find that Gaussian noise and quantization techniques can mitigate the privacy risks posed by Vec2Text, with quantization offering a simpler and more widely applicable solution. Our findings emphasize the need for caution in using text embeddings and highlight the importance of further research into robust defense mechanisms for NLP systems.
Our code and experiment results are available at \url{https://github.com/dqmis/vec2text-repro}.
\end{abstract}


\maketitle

\section{Introduction}

\begin{figure}[t]
\centering
 \includegraphics[width=1\linewidth]{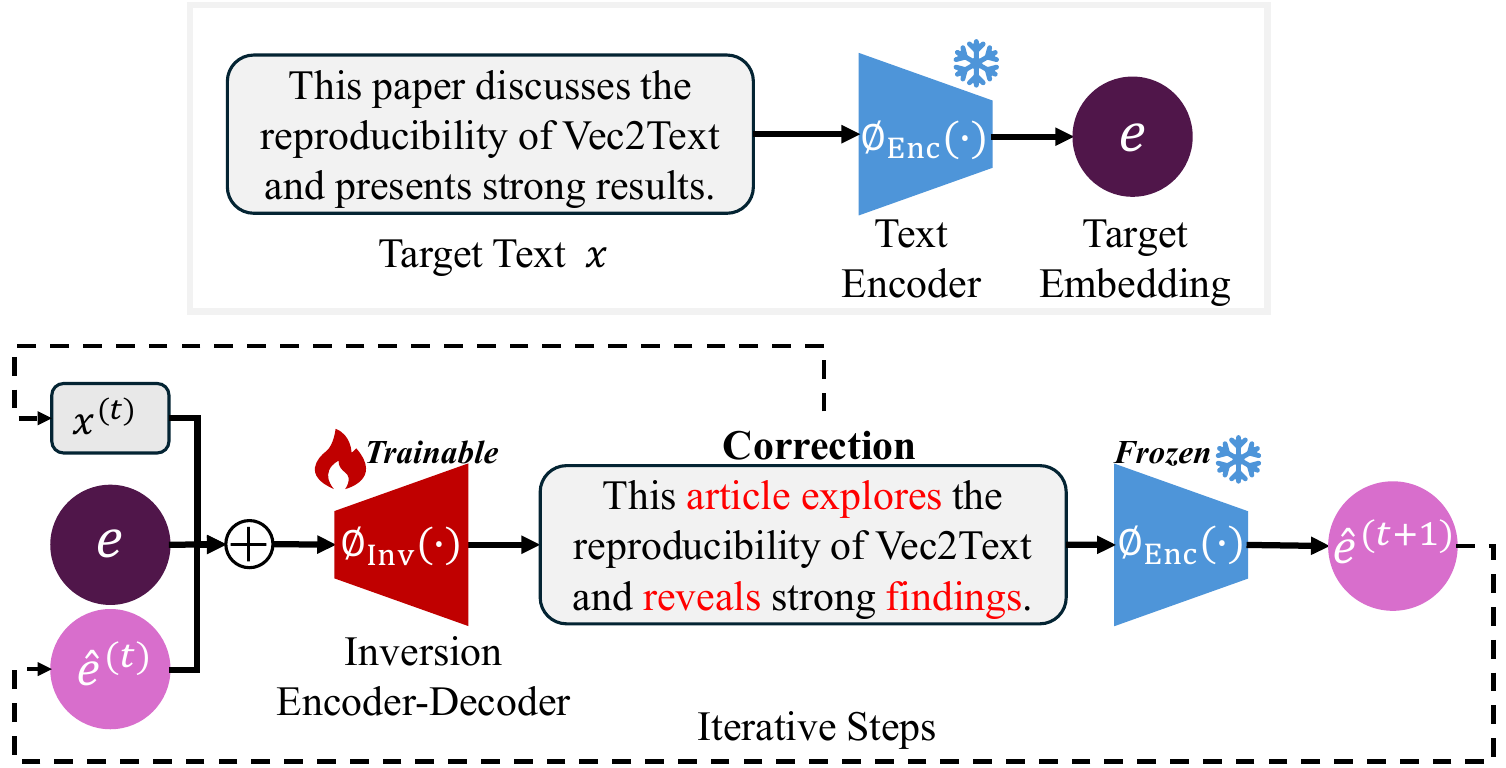}
\caption{The framework of Vec2Text reconstructs the target text $x$ from embeddings $e$ generated by a black-box encoder $\phi_{\text{Enc}}(\cdot)$. We train an inversion encoder-decoder-based model $\phi_{\text{Inv}}(\cdot)$ that iteratively generates a corrected text $x^{(t)}$ and aims to minimize the difference between $\hat{e}^{(t+1)}$
and the target embedding  $e$.
 $x^{(t)}$ and $\hat{e}^{(t+1)}$ are the input for the next iteration until the predefined number of steps is completed.\looseness=-1
}
\label{fig:vec2text}
\end{figure}

Text embeddings have become a cornerstone of modern NLP, enabling various applications such as machine translation~\cite{acl_MathurBC19,emnlp_MiSWI16,emnlp_XuDM21}, information retrieval~\cite{KarpukhinOMLWEC20_DPR,KhattabZ20_ColBERT,WangYHJYJMW23_SimLM}, and recommendation systems~\cite{kdd_ZhangYLXM16,kdd_OkuraTOT17,sigir_WuWQ021}. 
By mapping text into continuous vector spaces~\cite{naaclDevlinCLT19_BERT,emnlp_PenningtonSM14,Word2Vec}, embeddings capture intricate semantic relationships between linguistic units, allowing models to better understand and process language. \looseness=-1
Given the powerful capabilities of text embeddings, it has been recognized that a more privacy-preserving approach for third-party service providers is to collect only embeddings from the user side~\cite{icml_LeM14,nips_KirosZSZUTF15} and store them in their large-scale vector databases~\cite{douze2024faiss,qdrant2023}, rather than transmitting the raw text.

Therefore, a critical question arises: Can an attacker reconstruct the original text from embeddings, especially when it comes to high-value information such as passwords and personal privacy data? 
While this issue has been somewhat addressed in the computer vision domain~\cite{cvpr_DosovitskiyB16,cvpr_MahendranV15}, the exploration in the context of the text remains insufficient and less effective due to the gap between discrete tokens and continuous embeddings~\cite{song2020informationleakageembeddingmodels}.

Recently, Morris et al.~\cite{morris2023text} proposed a novel framework, Vec2Text, which trains a generator to iteratively approximate the text embedding during inference, ultimately recovering the original text (as shown in Figure~\ref{fig:vec2text}). 
Their approach demonstrates that, even with a black-box encoder model, the trained model achieves a  BiLingual Evaluation Understudy (BLEU)~\cite{PapineniRWZ02_BLEU} score of up to 97.3 for recovering 32-token texts compared to the original. 
The unexpectedly strong results, especially given that the latest method, GEIA~\cite{li2023sentenceembeddingleaksinformation}, only achieved a score of 1.0, motivated further verification, considering the typically non-intuitive and opaque structure of high-dimensional embedding spaces.
Additionally, we note that Zhuang et al.~\cite{zhuang2024understandingmitigatingthreatvec2text} attempted to reproduce Vec2Text's effectiveness in dense retrieval; however, they conducted all experiments using only the Natural Questions (NQ) dataset~\cite{KwiatkowskiPRCP19_NQ}, which limits the generalizability of their findings, particularly regarding training factors such as distance metrics, pooling functions, and embedding dimensions.\looseness=-1

Thus, our contributions in this reproducibility study are organized into two main aspects to achieve a comprehensive understanding of Vec2Text: (1) validating the original claims and (2) extending the study through targeted experiments.

First, we verify the performance claims of Vec2Text by reproducing experiments on both in-domain and out-of-domain datasets. 
We then evaluate Vec2Text's robustness against embedding inversion by testing its defense mechanism under additive Gaussian noise. 
Overall, we reproduce the results claimed in the original Vec2Text paper, demonstrating that Vec2Text is effective across multiple datasets.
However, some experiments yielded diverging results, which we suspect are due to differences in experimental details, such as model checkpoints and dataset splits. 
We also find that Vec2Text is sensitive to the length of the input text, which can easily lead to performance degradation.

Second, we conduct the following targeted extension experiments to understand Vec2Text from multiple perspectives:

(1) \textbf{Parameter Sensitivity Extension:} We examine how variations in the number of iterative steps and beam size influence text reconstruction performance. Our results provide a Pareto front, guiding the selection of optimal hyperparameter combinations to achieve the best trade-off between computational cost and reconstruction performance.\looseness=-1

(2) \textbf{Password Reconstruction Extension:}    While the original Vec2Text study evaluated sensitive text reconstruction using sentences from medical information, we assess its ability to reconstruct password texts, which are difficult to analyze for accurate semantic meaning due to their complex composition~\cite{ndss_VerasCT14,SE_PCFG}, pose a unique challenge for text reconstruction tasks. Our experiments show that, even in this more difficult scenario, Vec2Text can successfully invert a fraction of hard passwords.

(3) \textbf{Embedding Quantization Extension:} 
We evaluate Naive 8-bit quantization and Zeropoint quantization as defense mechanisms. While our quantization methods differ from those used by Zhuang et al.~\cite{zhuang2024understandingmitigatingthreatvec2text}, we similarly find that quantization techniques effectively reduce the reconstruction capability of Vec2Text while preserving retrieval performance. These findings highlight the general potential of embedding quantization methods to mitigate privacy risks associated with embedding inversion attacks.

In conclusion, our work not only reproduces the results claimed by Vec2Text but also extends the evaluation to new scenarios, providing further insights into its performance and robustness. Through these experiments, we demonstrate the strengths and limitations of Vec2Text in practical applications, contributing to the understanding of embedding inversion and its potential defenses. 
Our findings underscore the need for heightened vigilance in deploying text embeddings and advocate for continued research into robust, practical defense strategies to enhance both security and performance in future NLP and recommendation system applications. \looseness=-1

\section{Background and Related Work}

In this section, we review recent advances in text embeddings and embedding inversion techniques, along with their potential risks.\looseness=-1

\subsection{Text Embeddings}

To enable more effective natural language understanding by computer systems, researchers have developed embedding techniques to represent raw text numerically. The simplest approach employs one-hot encoding, where each word is represented as a sparse binary vector. Subsequent advancements introduced distributed representations through Word2Vec~\citep{Word2Vec}, which utilizes either: (1) the continuous bag-of-words (CBOW) architecture to predict target words from context, or (2) the skip-gram objective to predict context words from targets.
Building on these foundations, \citet{emnlp_PenningtonSM14} proposed GloVe, which constructs embeddings through factorization of global word-word co-occurrence matrices using weighted least-squares optimization. However, such bag-of-words approaches demonstrate limited capacity for capturing full contextual semantic context.
Modern embedding methods leverage deep Transformer architectures to address this limitation. For example, BERT \citep{naaclDevlinCLT19_BERT} employs bidirectional encoder representations through masked language modeling and next-sentence prediction objectives, enabling dynamic, context-aware token representations. With the rise of large language models (LLMs), recent work has explored decoder-only architectures for generating embeddings, as exemplified by RepLLaMA~\cite{MaWYWL24_RankLLaMA}, LENS~\cite{Yibing_LENS}, and MetaEOL~\cite{Yibin_MetaEOL}. These embedding techniques support diverse downstream applications, such as machine translation~\cite{acl_MathurBC19,emnlp_MiSWI16,emnlp_XuDM21}, information retrieval~\cite{KarpukhinOMLWEC20_DPR,KhattabZ20_ColBERT,WangYHJYJMW23_SimLM,li2025reproducinghotflip}, and recommendation systems~\cite{kdd_ZhangYLXM16,kdd_OkuraTOT17,sigir_WuWQ021}.

\subsection{Text Embedding Inversion}

Although significant progress has been made in recovering images~\cite{cvpr_DosovitskiyB16,cvpr_MahendranV15} from embeddings, reconstructing complete text from embeddings remains a challenging problem. The inherent gap between discrete tokens in text and the continuous nature of embedding space complicates this task. Bridging the gap between embedding space and token space not only deepens our understanding of embeddings but also facilitates efficient attack methods that can ultimately enhance model robustness. 
Research by \citet{song2020informationleakageembeddingmodels} initially demonstrates the feasibility of recovering textual information in the form of bags of words from embeddings. Subsequent studies \cite{info:doi/10.2196/18055, lehman2021doesbertpretrainedclinical} expose significant privacy risks by showing that sensitive data can be extracted from text embeddings. The Canary Extraction Attack (CEA) exemplifies this by extracting specific 'canary' tokens through prefix-driven text completion that iteratively predicts sensitive continuations of seeded phrases \cite{parikh2022canaryextractionnaturallanguage}. Similarly, the Generative Embedding Inversion Attack (GEIA) learns a mapping from a target embedding to an embedding that serves as a prompt to a decoder large language model \cite{li2023sentenceembeddingleaksinformation}. Recently, \citet{li2025unsupervisedcorpuspoisoningattacks} employ a classifier to recover token-level embeddings, which are then utilized for corpus poisoning attacks in information retrieval. In contrast, Morris et al.'s Vec2Text addresses the limitation of previous methods that could not reconstruct text in a single step by employing a multi-step iterative approach. Although this method increases inference time, its efficiency is significantly improved.\looseness=-1

\subsection{Limitations of Previous Work}
While the original Vec2Text~\cite{morris2023text} paper provides a compelling proof-of-concept for the embedding inversion method, it also has notable limitations that have inspired a range of follow-up studies. A reproducibility study by \citet{zhuang2024understandingmitigatingthreatvec2text} explores various factors, such as the influence of distance metrics, pooling functions, bottleneck pre-training, noise addition during training, embedding quantization, and embedding dimensions, on the performance of Vec2Text. 
Additionally, the application of Vec2Text to retrieval corpus poisoning is examined \cite{zhuang2024doesvec2textposenew}. In \citet{chen2024textembeddinginversionsecurity}, Vec2Text is extended to a multilingual setting with ad hoc translation and masking defense mechanisms. 
However, despite these advancements, the original Vec2Text method still faces challenges related to scalability, robustness, and sensitivity to varying input data characteristics, which continue to motivate further research. 
In this paper, while confirming several claims from the original work, we primarily focus on exploring the model's robustness to varying input text lengths. Additionally, we investigate the Pareto optimization of hyperparameters to minimize time costs while maximizing efficiency. Furthermore, we propose a quantization-based approach that effectively mitigates the privacy risks posed by Vec2Text.

\section{Reproducibility Methodology}

In this section, we first introduce the fundamentals of text embedding inversion and provide a formal definition of the problem. We then describe the Vec2Text framework proposed by Morris et al. and a method to mitigate potential text inversion attacks by introducing noise. These methods are the cornerstone of this reproduction work, and we will conduct experiments based on them.\looseness=-1

\subsection{Problem Formulation}  
Given a target text $x$ and an arbitrary text encoder \(\phi_{\text{Enc}}(\cdot)\). The encoder maps the text $x$  from discrete lexical space to continuous vector space and outputs its embedding as \(e = \phi_{\text{Enc}}(x)\).  Text embedding inversion refers to the process of finding an inversion model \(\phi_{\text{Inv}}(\cdot)\)  that generates a text $\hat{x}= \phi_{\text{Inv}}(e)$ from the embedding $e$, such that \(\hat{x}\) is approximately equal to the original text $x$, i.e., \(\hat{x}\simeq x\).

\subsection{The Vec2Text Method}

 As illustrated in Figure \ref{fig:vec2text}, the Vec2Text~\cite{morris2023text} employs a standard encoder-decoder model as the base model for training and transforms the text generation process into a multi-step procedure that iteratively corrects errors.

At each step, the input is divided into three parts:
\begin{enumerate}
    \item The target embedding $e$.
    \item The embedding of the text \({x}^{(t)}\) generated in the previous $t_{\text{th}}$ step \(\hat{e}^{(t)}\).
    \item Correction: The difference between the generated text's embedding and the target embedding: \(e-\hat{e}^{(t)}\). And the word-level embeddings \(\{w_1,w_2,...,w_n\}\) of the text generated in the previous $t_{\text{th}}$step, where $n$ is the number of tokens.
\end{enumerate}

The final input is their concatenation, as shown in the following:
\begin{align}
\text{Input} = \text{concat}\big( &{\text{EmbToSeq}}(e), \notag\\
                     &{\text{EmbToSeq}}(\hat{e}^{(t)}), \\
                     &{\text{EmbToSeq}}(e-\hat{e}^{(t)}),\left(w_{1}, w_{2},\ldots, w_{n}\right) \big)\notag
\end{align}
where \(\text{EmbToSeq}(e) = W_{2} \sigma (W_{1} e)\)  is a learnable Multi-Layer Perceptron~(MLP) to project vectors, and $W_{1}$ and $W_{2}$ are weight matrices and $\sigma$ is a non-linear activation function. Notably, for the three types of individual input vectors mentioned above, their respective \(\text{EmbToSeq}(\cdot)\) mappings are handled separately.

During training, we employ a standard language modeling loss. During inference, we perform multi-step inference iteratively and enhance model output by conducting greedy inference at the token level while implementing beam search at the sequence level.

\subsection{Defense Against Text Inversion}\label{sec:Defense Against Text Inversion}

Morris et al. also propose a method to defend against inversion attacks by adding Gaussian noise to the embeddings. This method is formally defined as:
\begin{equation}
    \phi_{\text{noisy}}(x) = \phi_{\text{Enc}}(x) + \lambda \cdot \epsilon, \quad \epsilon \sim \mathcal{N}(0,1)
\end{equation}
where $\lambda$ is a hyperparameter that scales the noise added to the embeddings. The authors argue that while small noise additions should not greatly decrease retrieval performance, they would effectively prevent the proposed inversion method from functioning correctly.\looseness=-1

\section{Experimental Setup}

In this section, we introduce the experimental setup, focusing on the datasets, models, evaluation metrics, and other components involved in the reproducibility experiments and the multiple extension experiments we designed.

\subsection{Datasets}

Following Morris et al.~\cite{morris2023text}, we use the same datasets the authors employed to evaluate their proposed method. Where applicable, we adopt the identical train, test, and validation splits defined by the authors. 
More specifically, we used two well-known datasets, Natural Questions (\textbf{NQ})~\cite{KwiatkowskiPRCP19_NQ} and \textbf{MSMARCO} ~\cite{msmarco}, to reproduce the in-domain reconstruction results from Morris et al. In addition, we conducted out-of-domain evaluation and extension experiments using datasets from the BEIR benchmark~\cite{thakur2021beir}, including \textbf{Quora}~\cite{quora2017dataset}, \textbf{Signal1M}~\cite{SuarezACME18_signal1m}, \textbf{Climate-Fever}~\cite{abs-2012-00614_climate_fever}, \textbf{Robust04}~\cite{Voorhees03b_robust04}, \textbf{BioASQ}~\cite{TsatsaronisBMPZ15_Bioasq}, \textbf{TREC-News}~\cite{SoboroffHH18_trec_news}, \textbf{ArguAna}~\cite{WachsmuthSS18_arguana}, \textbf{FiQA}~\cite{MaiaHFDMZB18_fiqa}, \textbf{NFCorpus}~\cite{BotevaGSR16_nfcorpus}, and \textbf{SciFact}~\cite{WaddenLLWZCH20_scifact}. Due to space limitations, further details about these datasets can be found in ~\cite{thakur2021beir}.

We also use a \textbf{Password Strength}~\cite{bhavik2019password} dataset to evaluate the efficiency of Vec2Text when processing password-format text. This dataset is derived from the 000Webhost password leak that occurred in October 2015. 
``\textit{lamborghin1}'' is a typical example in this dataset.

For the text embedding inversion task we focused on in this paper, we utilize only the corpus, i.e., the text portion, from all datasets. We also observe that Morris et al. do not use the entire corpus of the datasets but instead perform sampling for testing. Furthermore, we find that the sample size for these samples varies depending on the dataset and experiment. By investigating the experiment configurations provided in the code repository, we identify that the sample size for experimental evaluation ranges between 90 and 1,000 samples. Therefore, in this paper, when discussing each experiment, we will specify the exact sample size setting.

\subsection{Models}
\subsubsection{Text Encoders}

Morris et al. employ two text encoders as targets for training the text inversion models: the GTR-base encoder \cite{ni2021large}, a T5-based pre-trained transformer designed for text retrieval, and \textit{text-embeddings-ada-002} \cite{adaopen}, which is accessible via the OpenAI API.

\subsubsection{Inversion Models}\label{sec:Inversion Models}

The text inversion model is initialized
from the T5-base checkpoint~\cite{RaffelSRLNMZLL20_t5}.
Including the projection head, each model has approximately 235M parameters. The GTR-based inversion model is trained on the NQ  dataset with a maximum sequence length of 32 tokens (denoted as \textit{\textbf{gtr-nq-32}}). 
Morris et al.~\cite{morris2023text} train the \textit{text-embeddings-ada-002} inversion model on the MSMARCO dataset, producing two versions with a maximum sequence length of either 32 tokens~(denoted as \textit{\textbf{ada-ms-32}}) or 128 tokens~(denoted as \textit{\textbf{ada-ms-128}}) per example. 
However, only the version with a maximum output length of 128 tokens, \textit{\textbf{ada-ms-128}}, has been publicly released, while the \textit{\textbf{ada-ms-32}} model remains unavailable.
More importantly, aside from explicitly distinguishing between the two models in the in-domain results, Morris et al. do not clarify which version is used in the other experiments. Therefore, to ensure consistency, we use only \textit{\textbf{ada-ms-128}} in all experiments in this paper.\looseness=-1

Upon reviewing Morris et al.'s code repository, we note that they use additional model checkpoints not provided, which may differ in training epochs and sample size.
In the Section~\ref{sec:Experimental Results}, we discuss the potential impact of these unprovided checkpoints on the reproducibility of the experiments.\looseness=-1

\subsection{Evaluation Metrics}

Following Morris et al.~\cite{morris2023text}, to measure the word-match between
the original and inverted text, we use \textbf{BLEU} score~(typically on a scale of 0 to 100)~\cite{PapineniRWZ02_BLEU}, a measure of n-gram similarities between the true and reconstructed text; \textbf{Token F1}, the multi-class F1 score between the set of predicted tokens and the set of true tokens; \textbf{Exact-match}, the percentage of reconstructed outputs that perfectly match the ground truth. 
For the embedding similarity, we calculate the \textbf{Cosine} similarity between the target embedding and the embedding of reconstructed text through the text encoder \(\phi_{\text{Enc}}(\cdot)\).
The authors also employ the Normalized Discounted Cumulative Gain~(\textbf{nDCG@10}) metric for additional text retrieval experiments.

\subsection{Implementation Details}

The original codebase for the Vec2Text ~\cite{morris2023text} is publicly available and well-documented on GitHub\footnote{\url{https://github.com/vec2text/vec2text}}. To ensure consistency with their implementation of the Vec2Text method, we adhere to the methodology outlined in their work, using the code provided by the authors. We do not train our inversion models; instead, we utilize the provided weights for the reproducibility experiments. This approach guarantees alignment with the original study and facilitates a direct comparison of our results with those reported by the authors. \looseness=-1

Although the code is well-structured and designed for text inversion tasks, the authors do not provide replication scripts for most of the experiments testing their hypotheses. As a result, we lack crucial experimental details, including the exact models used, sample sizes, and hyperparameters. Moreover, we are unable to access pre-split datasets that would ensure consistent replication of the experiments. Finally, the absence of the \textit{\textbf{ada-ms-32}} checkpoint, which is essential for reproducing some experiments, presents significant challenges.
To replicate the experiments, we manually analyze the generated experimental outputs\footnote{\url{https://github.com/vec2text/vec2text/tree/proj/results\_evaluation}} and use reverse engineering to determine the possible hyperparameters used.
We also access OpenAI's API for experiments that require its use.
All our experiments are mainly implemented using Pytorch 2.1 on a Ubuntu server with NVIDIA H100 ×80G × 4 GPUs, AMD EPYC 9334 × 2 CPUs, and 768G memory.
We make our code, including all the experiment setups and necessary components, publicly available in a GitHub repository\footnote{\url{https://github.com/dqmis/vec2text-repro}}, to facilitate the replication of our work.

\section{Experimental Results}\label{sec:Experimental Results}

In this section, we present all the experimental results.  To provide a clear comparison and analysis, we divide the results into two parts: the reproduction of the experiments from the original paper and the extension experiments we designed.

\subsection{Reproducing Original Paper Results}
In this subsection, we conduct a series of detailed experiments to validate the four claims made in Vec2Text, evaluating its performance in in-domain and out-of-domain scenarios, as well as its robustness with Gaussian noise defense.

\begin{table*}[t]
\centering
\caption{\label{tab:tab1-repro}
In-domain reproduction performance of Vec2Text. Bold indicates reproduced results differing by over 10\% from Morris et al.
For the experiments with the MSMARCO corpus truncated to 32 tokens, Morris et al. use the \textit{\textbf{ada-ms-32}} model; however, since this model is not publicly available, we use the \textit{\textbf{ada-ms-128}} model to reproduce the experiments.}
\renewcommand\arraystretch{0.95} 
\setlength{\tabcolsep}{2.5mm}{
\begin{tabular}{@{}lcccccccccc@{}}
\toprule
\multicolumn{11}{c}{GTR - NQ ( Evaluate \textit{\textbf{gtr-nq-32}}, where the corpus is truncated to 32 tokens.)}                                                                                                                            \\ \midrule
\multirow{2}{*}{Method} & \multirow{2}{*}{\makecell[cc]{Token \\ Length}} & \multirow{2}{*}{\makecell[cc]{Pred Token \\ Length}}                                                  & \multicolumn{2}{c}{BLEU\(\,\uparrow\)}                            & \multicolumn{2}{c}{Token F1\(\,\uparrow\)}                          & \multicolumn{2}{c}{Exact-match\(\,\uparrow\)}                         & \multicolumn{2}{c}{Cosine\(\,\uparrow\)}       \\ \cmidrule(lr){4-5}\cmidrule(lr){6-7}\cmidrule(lr){8-9}\cmidrule(lr){10-11}  &&  & Original       & Ours & Original         & Ours & Original         & Ours & Original         & Ours 
\\ \midrule
\multicolumn{1}{l}{Base {[}0 steps{]}}     & 32     &{32}   & 31.9          &{31.4}           & 67          &{67}            & 0.0             &{0.0}            & 0.91          & 0.91          \\
\multicolumn{1}{l}{\qquad(+beam search)}        & 32     &{32}   & 34.5          &{34.4}           & 67          &{66}            & 1.0             &{1.0}            & 0.92          & 0.90           \\
\multicolumn{1}{l}{\qquad(+nucleus)}            & 32     &{32}   & 25.3          &{25.7}           & 60          &{61}            & 0.0             &{0.0}            & 0.88          & 0.91           \\
\multicolumn{1}{l}{Vec2Text {[}1 step{]}}  & 32     &{32}   & 50.7          &{49.4}           & 80          &{79}            & 0.0    &{\textbf{1.0}}   & 0.96          & 0.94          \\
\multicolumn{1}{l}{\qquad{[}20 steps{]}}         & 32     &{32}   & 83.9          &{83.6}           & 96          &{95}            & 40.2 &{\textbf{58.2}}  & 0.99          & 0.98          \\
\multicolumn{1}{l}{\qquad{[}50 steps{]}}         & 32     &{32}   & 85.4          &{84.6}           & 97          &{96}            & 40.6 &{\textbf{59.1}}  & 0.99          & 0.99          \\
\multicolumn{1}{l}{\qquad{[}50 steps + sbeam{]}} & 32     &{32}   & 97.3          &{98.5}           & 99          &{99}            & 92.0            &{94.0}           & 0.99          & 0.99          \\ \midrule
\multicolumn{11}{c}{OpenAI - MSMARCO ( Evaluate \textit{\textbf{ada-ms-128}}, where the corpus is truncated to 32 tokens.)}                                                                                                                                                                                                 \\ \midrule
\multirow{2}{*}{Method} & \multirow{2}{*}{\makecell[cc]{Token \\ Length}} & \multirow{2}{*}{\makecell[cc]{Pred Token \\ Length}}                                                  & \multicolumn{2}{c}{BLEU\(\,\uparrow\)}                            & \multicolumn{2}{c}{Token F1\(\,\uparrow\)}                          & \multicolumn{2}{c}{Exact-match\(\,\uparrow\)}                         & \multicolumn{2}{c}{Cosine\(\,\uparrow\)}       \\ \cmidrule(lr){4-5}\cmidrule(lr){6-7}\cmidrule(lr){8-9}\cmidrule(lr){10-11}  &&  & Original       & Ours & Original         & Ours & Original         & Ours & Original         & Ours 
\\ \midrule
\multicolumn{1}{l}{Base {[}0 steps{]}}     & 31.8   &{31.8} & 26.2 &{\textbf{20.8}}  & 61 &{\textbf{53}} & 0.0            &{0.0}            & 0.94 & 0.88 \\
\multicolumn{1}{l}{Vec2Text {[}1 step{]}}  & 31.8   &{31.9} & 44.1          &{\textbf{40.8}}           & 77          &71            & 5.2  &{\textbf{2.6}}   & 0.96 & 0.91 \\
\multicolumn{1}{l}{\qquad{[}20 steps{]}}         & 31.8   &{31.9} & 61.9 &{\textbf{37.8}} & 87 &{\textbf{67}} & 15.0   &{\textbf{2.3}} & 0.98 & 0.91 \\
\multicolumn{1}{l}{\qquad{[}50 steps{]}}         & 31.8   &{31.9} & 62.3 &{\textbf{37.9}}  & 87 &{\textbf{68}}   & 14.8 &{\textbf{3.4}}   & 0.98 & 0.92 \\
\multicolumn{1}{l}{\qquad{[}50 steps + sbeam{]}} & 31.8   &{31.8} & 83.4 &{\textbf{43.4}}  & 96 &{\textbf{74}}   & 60.9          &{\textbf{4.8}}            & 0.99 & 0.92 \\ \midrule
\multicolumn{11}{c}{OpenAI - MSMARCO (  Evaluate \textit{\textbf{ada-ms-128}}, where the corpus is truncated to 81 tokens. )}                                                                                                                                                                                                \\ \midrule
\multirow{2}{*}{Method} & \multirow{2}{*}{\makecell[cc]{Token \\ Length}} & \multirow{2}{*}{\makecell[cc]{Pred Token \\ Length}}                                                  & \multicolumn{2}{c}{BLEU\(\,\uparrow\)}                            & \multicolumn{2}{c}{Token F1\(\,\uparrow\)}                          & \multicolumn{2}{c}{Exact-match\(\,\uparrow\)}                         & \multicolumn{2}{c}{Cosine\(\,\uparrow\)}       \\ \cmidrule(lr){4-5}\cmidrule(lr){6-7}\cmidrule(lr){8-9}\cmidrule(lr){10-11}  &&  & Original       & Ours & Original         & Ours & Original         & Ours & Original         & Ours 
\\ \midrule
\multicolumn{1}{l}{Base {[}0 steps{]}}     & 80.9   &{84.2} & 17.0            &{17.4}           & 54          &{54}            & 0.6           &{0.6}            & 0.95          & 0.96          \\
\multicolumn{1}{l}{Vec2Text {[}1 step{]}}  & 80.9   &{81.6} & 29.9          &{32.2}           & 68          &{69}            & 1.4           &{1.4}            & 0.97          & 0.97          \\
\multicolumn{1}{l}{\qquad{[}20 steps{]}}         & 80.9   &{79.7} & 43.1          &{44.7}           & 78          &{78}            & 3.2           &{3.1}            & 0.99          & 0.99          \\
\multicolumn{1}{l}{\qquad{[}50 steps{]}}         & 80.9   &{80.5} & 44.4          &{45.1}           & 78          &{79}            & 3.4           &{3.4}            & 0.99          & 0.99          \\
\multicolumn{1}{l}{\qquad{[}50 steps + sbeam{]}} & 80.9   &{80.6} & 55.0            &{54.3}           & 84          &{84}          & 8.0             & 8.1                               & 0.99          & 0.99          \\ \bottomrule
\end{tabular}}
\end{table*}

We report the experimental results in two types: 
\textbf{1)}~\textbf{Original}: The results reported by Morris et al. in their paper.
\textbf{2)}~\textbf{Ours}: The results obtained by running the experiments with the same setup.
Whenever possible, we report both sets of results to ensure comparability.\looseness=-1

\begin{figure}[!t]
\centering
\includegraphics[width=\linewidth]{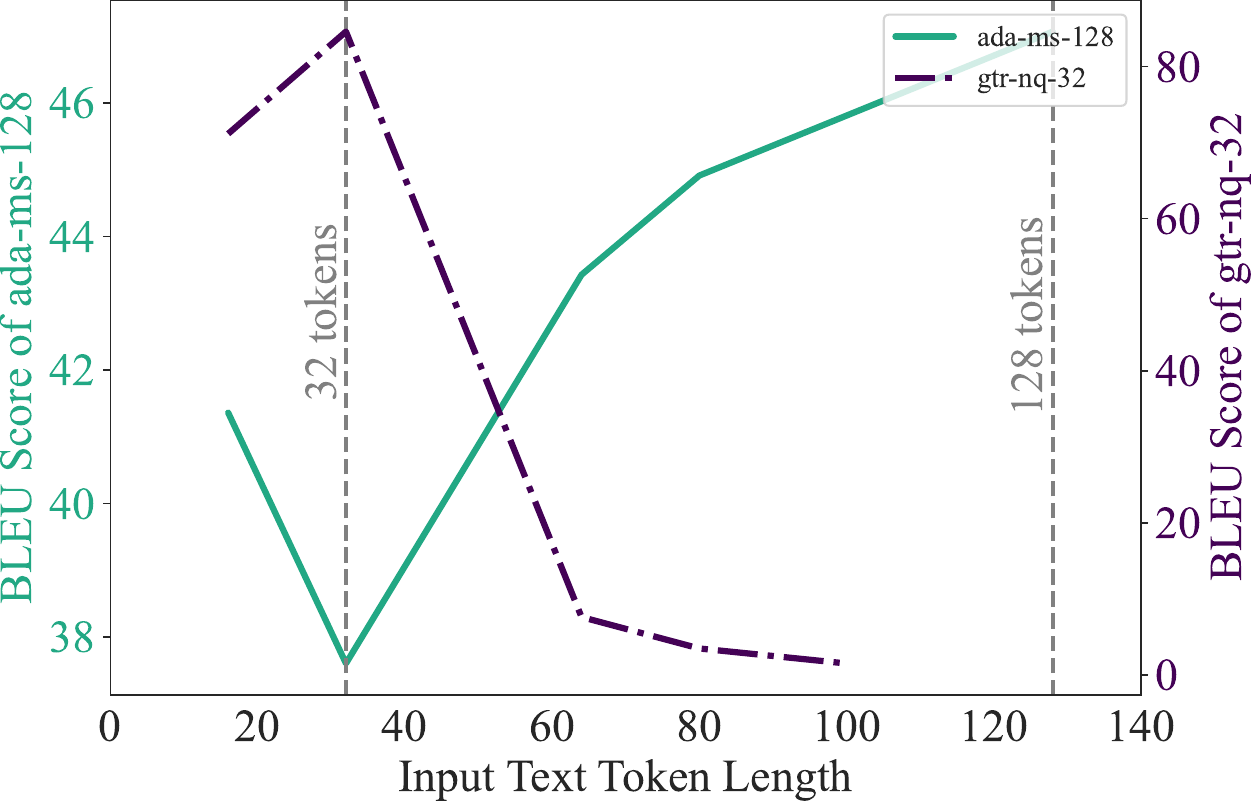}
\caption{The reconstruction performance of the \textit{gtr-nq-32} and \textit{ada-ms-128} models when given input texts of varying lengths. Their differences in performance curves likely stem from variations in their respective training objectives.
\looseness=-1}
\label{fig:seq len}
\end{figure}

\subsubsection{Claim 1 - Vec2Text can reconstruct in-domain text from given embeddings}\label{sec:claim1}
Here, ``in-domain'' refers to a scenario where the model is trained and evaluated on text reconstruction within the scope of a single dataset. We verify this capability by reproducing the in-domain reconstruction setting for two inversion models that the authors provided: \textit{gtr-nq-32} and \textit{ada-ms-128}, using the NQ and MSMARCO datasets. Following Morris et al.~\cite{morris2023text}, we utilize 1000 samples from each dataset and the same configuration.
The results of our experiments are reported in Table \ref{tab:tab1-repro}.

From Table \ref{tab:tab1-repro}, we observe that the \textit{gtr-nq-32} model, with the corpus truncated to 32 tokens, and the \textit{ada-ms-128} model, with the corpus truncated to 81 tokens, yield nearly identical performance across all evaluation metrics. Notably, the \textit{gtr-nq-32} model achieves a marginally higher exact match score when evaluated on the NQ dataset. Additionally, we observe significant differences in the performance when evaluating the \textit{ada-ms-128} model with the corpus truncated to 32 tokens, compared to the results described in the original paper. We hypothesize that this discrepancy arises because Morris et al. trained a separate model specifically for 32 tokens to invert the \textit{text-embeddings-ada-002} model. In our case, due to the lack of this model, we were compelled to use the 128-token version \textit{ada-ms-128} for testing. This point is also discussed in Section~\ref{sec:Inversion Models}.

The aforementioned issue prompts us to investigate how well-trained inversion models perform when confronted with texts of varying lengths. We conducted tests using MSMARCO and compiled the BLEU scores for documents of different lengths, with the results presented in Figure~\ref{fig:seq len}.
From Figure~\ref{fig:seq len}, we observe that both \textit{gtr-nq-32} and \textit{ada-ms-128} achieve optimal performance when reconstructing texts whose lengths match their respective training text lengths—specifically, 32 and 128 tokens. However, a notable difference is that \textit{ada-ms-128} performs poorly at 32 tokens. We do not have a clear explanation for this observation but speculate that it is due to specific adjustments made by Morris et al. during the training of \textit{ada-ms-128}, which likely aimed to complement the unpublished 32-token length model.

Based on the above experiments, we confirm that the in-domain performance of Vec2Text is reproducible. However, we also find that the trained inversion models are sensitive to the length of the input text. Specifically, they exhibit degraded performance when applied to out-of-domain sequence lengths. This limitation suggests a potential direction for improving the Vec2Text method.

\subsubsection{Claim 2 - Vec2Text is able to reconstruct out-of-domain text from given embeddings}
Morris et al.~\cite{morris2023text} experimentally demonstrate the effectiveness of Vec2Text in out-of-domain settings. We reproduce this experiment in Table~\ref{tab:tab2-repro}. In particular, based on the performance variations observed in Section~\ref{sec:claim1} when models are presented with texts of different lengths, we select seven datasets from BEIR that span a range of text lengths and compare the results using both the MSMARCO and NQ datasets.

However, we face reproduction challenges due to an under-specified experimental setup: the inversion model used and key inference hyperparameters (such as beam size and the number of iterative steps) are not clearly stated. Through inspection of the code and logs, we determine that the authors employ the \textit{ada-ms-128} model with a beam size of 8 and 50 iterative steps. The number of evaluation samples varies between 90 and 200, and for consistency, we standardize our experiments using 200 samples.

\begin{table}[t]
\centering
\caption{\label{tab:tab2-repro} Out-of-domain reproduction performance. We bold our reproduced results when the change rate exceeds 10\% compared to the original results reported by Morris et al. The base method refers to the Vec2Text method without the use of feedback (0 iteration steps).}
\renewcommand\arraystretch{0.95} 
\setlength{\tabcolsep}{1mm}{
\begin{tabular}{@{}lrlrrrr@{}}
\toprule
\multirow{2}{*}{Dataset} &\multirow{2}{*}{\makecell[cc]{Token \\ Length}} & \multirow{2}{*}{Method}  & \multicolumn{2}{c}{BLEU\(\,\uparrow\)}      & \multicolumn{2}{c}{Token F1\(\,\uparrow\)}       \\ \cmidrule(lr){4-5} \cmidrule(lr){6-7} &&
 & Original         & Ours          & Original         & Ours          \\ \midrule
\multirow{2}{*}{Quora}         & \multirow{2}{*}{15.7}  & Base     & 36.2 & \textbf{40.0}   & 73.8 & 77.2   \\
                               &                        & Vec2Text & 95.5          & 95.1          & 98.6          & 98.5            \\ \midrule
\multirow{2}{*}{Signal1M}      & \multirow{2}{*}{23.7}  & Base     & 13.2          & \textbf{10.4}          & 49.5          & 45.6            \\
                               &                        & Vec2Text & 80.7 & \textbf{57.7} & 92.5 & \textbf{80.1} \\ \midrule
\multirow{2}{*}{MSMARCO}       & \multirow{2}{*}{72.1}  & Base     & 15.5          & \textbf{17.2}          & 54.1          & 55.0            \\
                               &                        & Vec2Text & 59.6          & 58.1          & 86.1          & 84.9            \\ \midrule
\multirow{2}{*}{Climate-Fever} & \multirow{2}{*}{73.4}  & Base     & 12.8          & 13.2          & 49.3          & 50.1          \\
                               &                        & Vec2Text & 44.9          & \textbf{54.8}          & 82.6          & 80.5            \\ \midrule
\multirow{2}{*}{NQ}            & \multirow{2}{*}{94.7}  & Base     & 11.0   & \textbf{6.2}  & 47.1 & \textbf{36.5}   \\
                               &                        & Vec2Text & 32.7 & \textbf{14.7} & 72.7 & \textbf{53.0}   \\ \midrule
\multirow{2}{*}{Robust04}      & \multirow{2}{*}{127.3} & Base     & 4.9           & 4.6           & 34.4          & 32.1            \\
                               &                        & Vec2Text & 15.5          & 14.8          & 54.5          & \textbf{43.9}            \\ \midrule
\multirow{2}{*}{BioASQ}        & \multirow{2}{*}{127.4} & Base     & 5.3           & \textbf{3.6}           & 35.7          & \textbf{30.6}            \\
                               &                        & Vec2Text & 22.8 & \textbf{8.6}  & 59.5 & \textbf{42.7}   \\ \midrule
\multirow{2}{*}{SciFact}       & \multirow{2}{*}{127.4} & Base     & 4.9           & \textbf{3.1}           & 35.7          & \textbf{30.2}            \\
                               &                        & Vec2Text & 16.6 & \textbf{9.14} & 59.5 & \textbf{43.8}   \\ \midrule
\multirow{2}{*}{TREC-News}     & \multirow{2}{*}{128.0}   & Base     & 4.9           & \textbf{3.9}           & 34.8          & \textbf{28.0}            \\
                               &                        & Vec2Text & 14.5 & \textbf{7.9}  & 51.5 & \textbf{36.4}   \\ \bottomrule
\end{tabular}
}
\end{table}

As shown in Table~\ref{tab:tab2-repro}, we observe that the out-of-domain performance of Vex2Text on several datasets (MSMARCO, Climate-Fever, and Robust04) aligns with the results reported by Morris et al. However, our reproduced results on other datasets are generally lower than those presented in the original paper. 
Upon investigating the potential causes of these discrepancies, we identify two issues by reviewing the experiment logs in the code repository: 1) Morris et al. may have used multiple versions of the \textit{ada-ms-128} model, which were trained with more iterative steps; and 2) Morris et al. did not standardize the sample size during evaluation, resulting in sample sizes as small as 90 for some datasets, whereas our experiment uses a standardized sample size of 200. These discrepancies in model versions and sample sizes introduce significant variability, which may affect the consistency of the results.

\subsubsection{Claim 3 - Gaussian noise can be used to defend against Vec2Text}\label{sec:Claim 3}
As outlined in Section~\ref{sec:Defense Against Text Inversion}, Morris et al.~\cite{morris2023text} propose a potential defense against Vec2Text by introducing Gaussian noise to the embedding in a manner that preserves its functionality (e.g., retrieval performance) while effectively rendering Vec2Text ineffective. To evaluate the effectiveness of this approach, Morris et al. employ the \textit{gtr-nq-32} model and select 15 datasets from the BEIR benchmark, including MSMARCO and NQ. They introduce noise to the embeddings at five different scales, $\lambda = \{0, 0.001, 0.01, 0.1, 1.0\}$, while assessing the retrieval performance and the reconstruction performance of Vec2Text (the sampling size remains undisclosed). 
\begin{figure}[t]
\centering
\includegraphics[width=\linewidth]{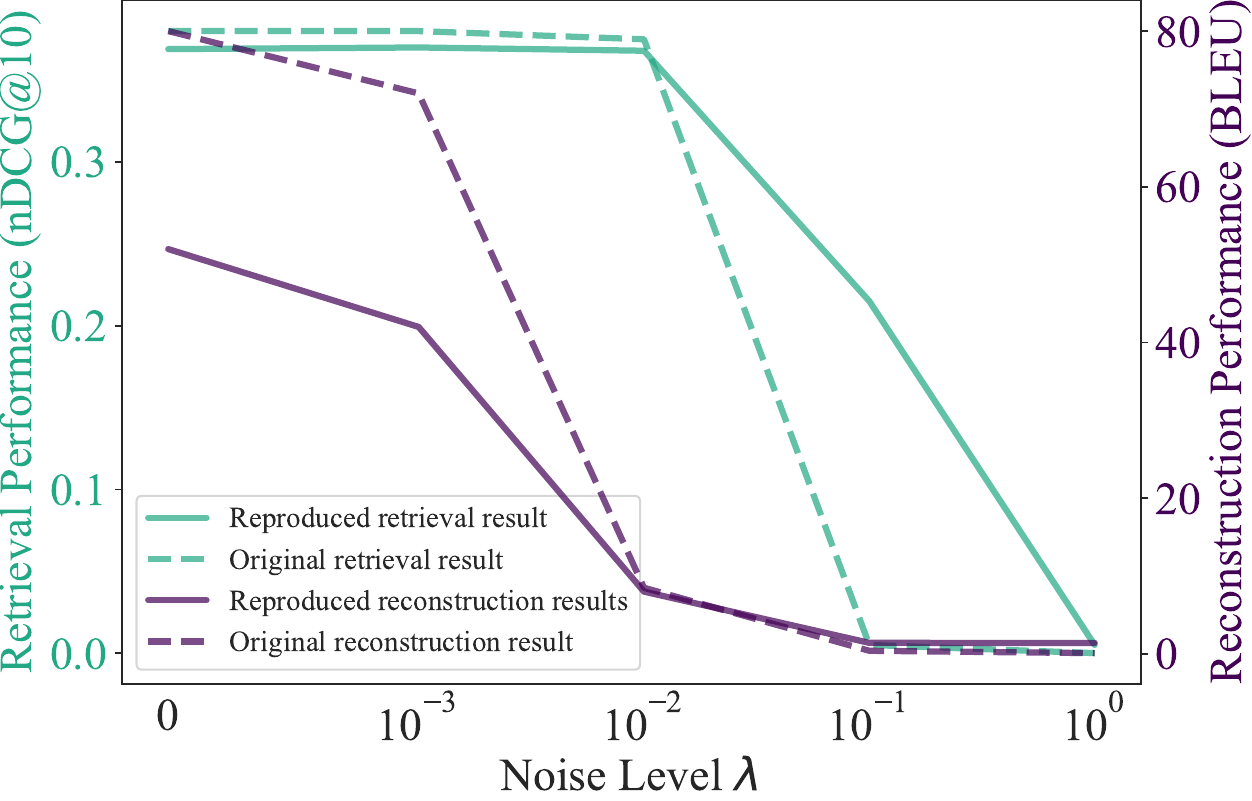}
\caption{Comparison of retrieval (nDCG@10) and reconstruction (BLEU) performance under varying levels of noise injection between the original study and our reproduction.}
\label{fig:fig2_comparision}
\end{figure}

We replicate this experiment; however, due to computational limitations, we select only five datasets from the original benchmark: ArguAna, FiQA, NFCorpus, Quora, and SciFact. For each dataset, we sample 1000 queries to evaluate retrieval performance and 1000 documents to assess the reconstruction performance of Vec2Text. 

We present the experimental results in Figure~\ref{fig:fig2_comparision}, following the format of Morris et al.~\cite{morris2023text}. As shown in Figure 3, we replicate their results in terms of overall trends, indicating that adding Gaussian noise effectively mitigates the risks posed by Vec2Text, with the best results observed when $\lambda=0.01$. We also find that the BLEU scores obtained in our experiment are lower than those of Morris et al., likely due to differences in dataset selection, as the data we chose exhibit lower scores in their out-of-domain tests.

In summary of our experiments, we find that the three claims in Vex2Text can be reproduced to some extent and demonstrate good performance. Concurrently, we observe that Vex2Text exhibits lower robustness in out-of-domain testing, and the model is sensitive to the length of the input text. This insight contributes to the analysis of more effective defense methods against potential privacy leakage issues posed by Vec2Text.

\subsection{Insights from Extension Experiments}

In this subsection, we design experiments to complete three extensions: the Parameter Sensitivity Extension, the Password Reconstruction Extension, and the Embedding Quantization Extension. Through these three extension experiments, we gain a more comprehensive understanding of Vec2Text.

\begin{figure}[t]
\centering
\includegraphics[width=\linewidth]{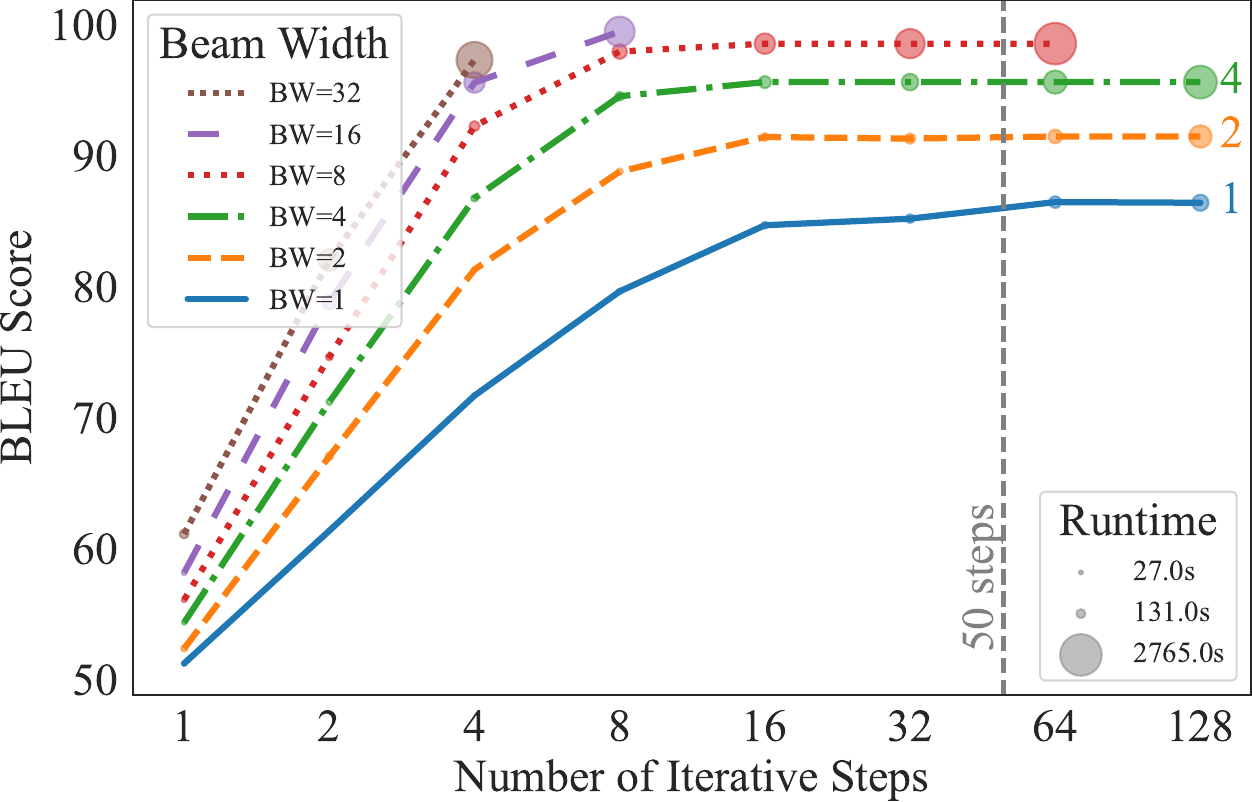}
\caption{Trends in BLEU score and runtime with varying iterative steps and beam width using the \textit{gtr-nq-32} model and the NQ dataset in Vec2Text.}
\label{fig:parameter_trade_off}
\end{figure}
\subsubsection{Parameter Sensitivity Extension}

Morris et al. observe that Vec2Text improves monotonically as the number of iterative steps and beam width increase. However, they do not conduct detailed experiments to explore the impact of these hyperparameters. Therefore, we test the performance of Vec2Text using the \textit{gtr-nq-32} model and the NQ dataset with beam widths of {1, 2, 4, 8, 16, 32} and iterative steps of {1, 2, 4, 8, 16, 32, 64, 128}. 
 The experimental results, as shown in Figure \ref{fig:parameter_trade_off}, reveal that both increasing the beam width and the number of iterative steps improve the performance. However, if we aim to control the runtime, prioritizing the increase in the number of iterative steps is more likely to lead to a performance boost.\looseness=-1
 
\begin{figure}[t]
\centering
\includegraphics[width=\linewidth]{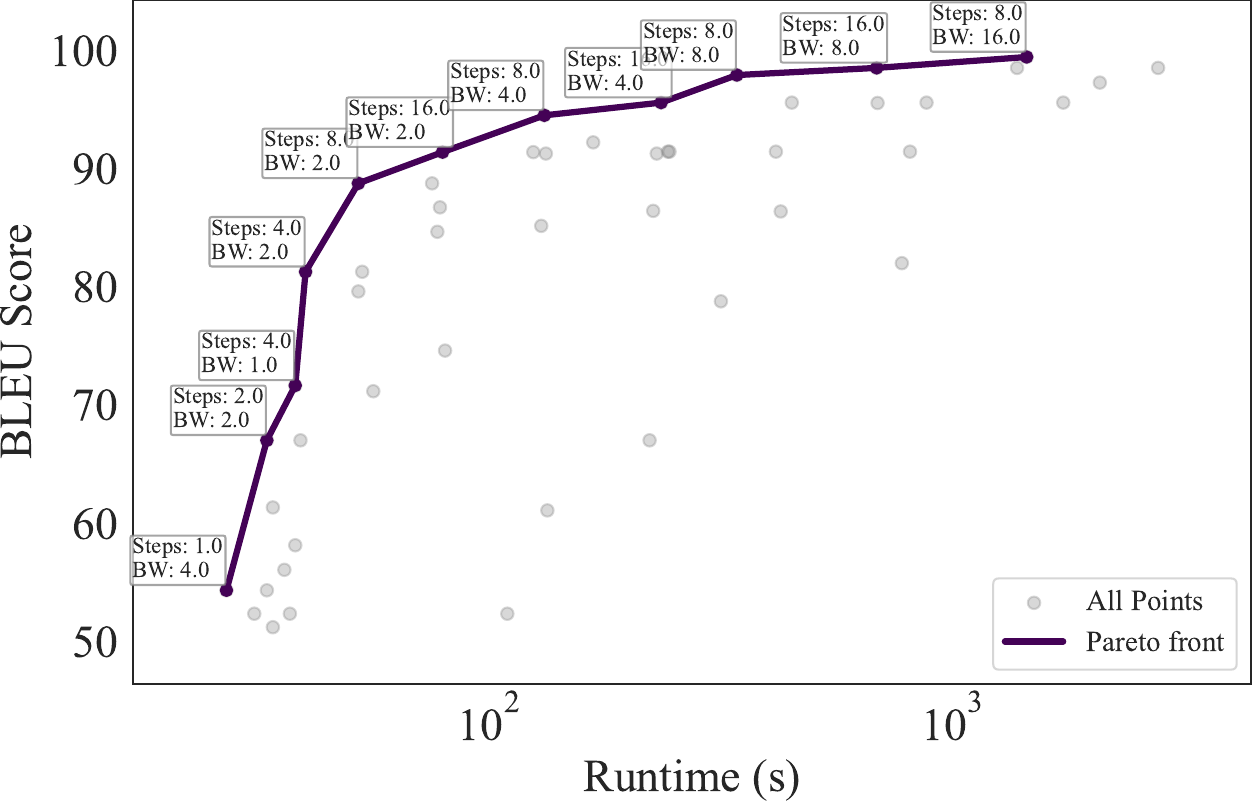}
\caption{Points on the Pareto front showing optimal trade-offs between BLEU score and runtime by choosing the number of iterative steps and beam width.\looseness=-1}
\label{fig:pareto_front}
\end{figure}

Another important consideration is the efficiency of the method, specifically its runtime. We aim to understand how the selection of beam width and iterative steps can maximize Vec2Text performance for a fixed runtime. 
Therefore, all data points from the aforementioned Figure \ref{fig:parameter_trade_off} are utilized to construct the Pareto front, with the resulting curve presented in Figure \ref{fig:pareto_front}. 
For any given runtime, Figure \ref{fig:pareto_front} can guide us in selecting the optimal combination of hyperparameters to achieve the optimal performance.

\begin{table}[t]
\centering
\captionof{table}{ Password examples from the Password Strength dataset, where passwords are classified into three strengths~(Easy, Medium, and Hard).
}
\renewcommand\arraystretch{1} 
\label{tab:passw_examples}
\begin{tabular}{@{}ll@{}}
\toprule
{Strength}           & \multicolumn{1}{c}{{Password}}               \\ \midrule
\multicolumn{1}{l}{Easy}   & josue12, mafia4, sandy13                \\
\multicolumn{1}{l}{Medium} & tribon1399, huevos1998, jdysfguda18             \\
\multicolumn{1}{l}{Hard}   & Nijiclanevolutionteam4, 5V7bWkzQxNAPVBzI \\
\bottomrule
\end{tabular}
\end{table}

\begin{table}[t]
\centering

\caption{\label{tab:password-results}Vec2Text performance on password reconstruction across different strength levels. BLEU was excluded due to its reliance on n-gram overlaps, which renders it statistically unreliable when the token length is low.
}
\renewcommand\arraystretch{0.85} 
\setlength{\tabcolsep}{0.6mm}{
\begin{tabular}{@{}lccccc@{}}
\toprule
 \multirow{2}{*}{Dataset} & \multirow{2}{*}{\makecell[cc]{Token \\ Length}}
& \multicolumn{2}{c}{\textbf{gtr-nq-32}} & \multicolumn{2}{c}{\textbf{ada-ms-128}} \\ \cmidrule(lr){3-4} \cmidrule(lr){5-6} 
 && {Exact-match} & \multicolumn{1}{c}{{Token F1}} & {Exact-match}  & {Token F1} \\ \midrule
\multicolumn{1}{l}{Easy}    & \multicolumn{1}{c}{3.5}              & 0              & \multicolumn{1}{c}{37}           & 36              & 36           \\
\multicolumn{1}{l}{Medium}  & \multicolumn{1}{c}{4.6}              & 0              & \multicolumn{1}{c}{21}           & 22              & 22           \\
\multicolumn{1}{l}{Hard}    & \multicolumn{1}{c}{11.2}             & 0              & \multicolumn{1}{c}{3}            & 4               & 6            \\ \bottomrule
\end{tabular}
}
\end{table}

\begin{table*}[t]
\centering
\caption{\label{tab:quant-res} Impact of different quantization methods on mitigating privacy risks from Vec2Text. Bold indicates quantized results differing by over 10\% from those without quantization. The goal of quantization is to reduce Vec2Text’s reconstruction performance (measured by BLEU) while preserving retrieval performance (measured by nDCG@10).
}
\renewcommand\arraystretch{0.95} 
\setlength{\tabcolsep}{1.1mm}{
\begin{tabular}{lcccccccccc} 
\toprule
\multirow{2}{*}{\makecell[cc]{Quantization\\Method}} & \multicolumn{2}{c}{ArguAna} & \multicolumn{2}{c}{FiQA} & \multicolumn{2}{c}{NFCorpus} & \multicolumn{2}{c}{Quora} & \multicolumn{2}{c}{SciFact}                             \\ 
\cmidrule(l){2-11}
                        & nDCG@10\(\,\uparrow\) & BLEU\(\,\uparrow\)              & nDCG@10\(\,\uparrow\) & BLEU\(\,\uparrow\)           & nDCG@10\(\,\uparrow\) & BLEU\(\,\uparrow\)               & nDCG@10\(\,\uparrow\) & BLEU\(\,\uparrow\)            & \multicolumn{1}{l}{nDCG@10\(\,\uparrow\)} & \multicolumn{1}{l}{BLEU\(\,\uparrow\)}  \\ 
\cmidrule{1-11}
w/o quantization         & 0.281   & 59.1              & 0.207   & 38.5           & 0.246   & 63.1               & 0.844   & 35.9            & 0.261                       & 56.6                      \\
Absolute Maximum             & 0.281   & \textbf{19.0}     & 0.206   & \textbf{16.0}  & 0.245   & \textbf{20.7}      & 0.844   & \textbf{27.3}   & 0.256                       & \textbf{24.2}             \\
Zeropoint               & 0.279   & \textbf{19.0}     & 0.208   & \textbf{16.0}  & 0.245   & \textbf{20.5}      & 0.844   & \textbf{27.4}   & 0.256                       & \textbf{24.8}             \\
\bottomrule
\end{tabular}}
\end{table*}

\subsubsection{Password Reconstruction Extension}
Morris et al. examine Vec2Text's ability to reconstruct sensitive information from sources such as the MIMIC III \cite{johnson2016mimic} medical dataset, where data typically possesses clear semantic content. Building on this, and to achieve a more challenging and comprehensive understanding of Vec2Text's performance when confronted with sensitive information that lacks explicit semantic structure, we focus on the task of password reconstruction.
We conduct experiments using the Password Strength~\cite{bhavik2019password} dataset and two models, \textit{gtr-nq-32} and \textit{ada-ms-128}, along with their corresponding encoders. The Password Strength dataset classifies passwords into three strengths: Easy, Medium, and Hard, with examples provided in Table \ref{tab:passw_examples}. We sample 1000 passwords from each difficulty level for the embedding inversion test. The experimental results are shown in Table \ref{tab:password-results}.\looseness=-1

From Table \ref{tab:password-results}, we observe that the \textit{gtr-nq-32} model cannot reconstruct passwords of any strength, despite having a token F1 score similar to that of \textit{ada-ms-128}. In contrast, the \textit{ada-ms-128} model successfully recovers 36\% of ``Easy'' passwords, 22\% of ``Medium'' level passwords, and 4\% of ``Hard'' passwords.
The reason for the poor performance of \textit{gtr-nq-32} remains unclear; it may be due to the smaller training dataset compared to \textit{ada-ms-128}. From the above experimental results, we can observe that despite \textit{ada-ms-128} not being fine-tuned for the password reconstruction task, it still yields effective results. This highlights the potential danger of embedding leakage, particularly with passwords used for online logins.

\subsubsection{Embedding Quantization Extension}
Morris et al. propose a method to mitigate the threats posed by Vec2Text by adding Gaussian noise to the embedding. As we have validated in Section~\ref{sec:Claim 3}, this approach is effective. However, it requires precise fine-tuning of the noise scale (i.e., the $\lambda$ parameter), as improper adjustment may easily negatively affect retrieval performance. 

Since quantization introduces discretization errors that disproportionately affect the fine-grained information needed to reconstruct the original inputs, while largely preserving the coarse-grained relationships essential for retrieval tasks, we build on this idea by exploring embedding quantization as a lightweight, noise-free alternative defense mechanism. Specifically, we apply 8-bit quantization to corpus embeddings, introducing small but structured perturbations that result in information loss.
We investigate two quantization techniques~\cite{weight_quantization, NEURIPS2022_8_bit_Quantization}: \textit{Absolute Maximum Quantization} and \textit{Zeropoint Quantization}, as described below.

\vspace{0.5em}
\noindent
\textbf{Absolute Maximum Quantization:}
\begin{align}
 \phi_{\text{q}_{\text{Enc}}}(x) &= \text{round}\left(\frac{127}{\max(\phi_{\text{Enc}}(x))} \cdot \phi_{\text{Enc}}(x)\right) \\
 \hat{\phi}_{\text{Enc}}(x) &= \frac{\max(\phi_{\text{Enc}}(x))}{127} \cdot \phi_{\text{q}_{\text{Enc}}}(x)
    \end{align}


\noindent
\textbf{Zeropoint Quantization:} 
\begin{align}
S &= \frac{255}{\max(\phi_{\text{Enc}}(x)) - \min(\phi_{\text{Enc}}(x))} \\
 Z &= - \text{round}(S \cdot \min(\phi_{\text{Enc}}(x))) - 128 \\
 \phi_{\text{q}_{\text{Enc}}}(x) &= \text{round}(S \cdot \phi_{\text{Enc}}(x) + Z) \\
\hat{\phi}_{\text{Enc}}(x) &= \frac{\phi_{\text{q}_{\text{Enc}}}(x) - Z}{S}
\end{align}
where $ \phi_{\text{q}_{\text{Enc}}}(\cdot)$ is the quantized encoding function, and $\hat{\phi}_{\text{Enc}}(\cdot)$ is used to restore the transmitted embedding back into an integer after reception, allowing it to be used for retrieval.

Following the setup in Section~\ref{sec:Claim 3}, we use the \textit{gtr-nq-32} model and five datasets from the BEIR benchmark: ArguAna, FiQA, NFCorpus, Quora, and SciFact. The experimental results shown in Table \ref{tab:quant-res} align with our theoretical analysis: while retrieval performance (measured by nDCG@10) remains stable, the inversion capability of Vec2Text (measured by BLEU) exhibits a significant drop. 
This contrast confirms that the proposed quantization method effectively preserves retrieval quality while mitigating privacy risks from embedding inversion attacks.

Unlike the Gaussian noise method in Section~\ref{sec:Claim 3}, which requires fine-tuning of the noise scale, quantization is a lightweight, hyperparameter-free method that can be rapidly deployed. 
However, it introduces a trade-off between utility and invertibility, as quantization does not remove all sensitive information from embeddings.
An adaptive adversary, aware of the quantization scheme, could potentially train quantization-aware inversion models or apply denoising techniques to partially recover the lost details. \looseness=-1

\section{Conclusion}

In this reproducibility study, we replicate three key claims from Vec2Text, confirming its strong performance and observing trends that align with those reported by Morris et al.~\cite{morris2023text}, though with some differences.
We observe that Vec2Text is sensitive to input text length, with significant degradation when it deviates from the training maximum. Additionally, adding Gaussian noise to embeddings effectively mitigates privacy risks, but requires precise fine-tuning of the scale parameter $\lambda$, as improper adjustment can significantly impair retrieval performance.

Moreover, we design three extensions for further insights into Vec2Text. The Parameter Sensitivity Extension provides a Pareto curve illustrating optimal hyperparameter combinations balancing time consumption and text reconstruction performance. The Password Reconstruction Extension demonstrates Vec2Text's capability to reconstruct passwords, even without explicit semantic structure. The Embedding Quantization Extension offers a simple, hyperparameter-free method to reduce privacy leakage, more deployable than the Gaussian noise method.

In conclusion, while Vec2Text is a powerful tool for model inversion, this study emphasizes the need for thorough experimental reporting and suggests future research on enhancing model robustness and mitigation strategies. Specifically, we believe that in embedding-based recommendation systems, reconstructing user behavior history poses a potential risk. Investigating how Vec2Text can work with user embeddings to reconstruct user history could be a promising direction for privacy-preserving research.\looseness=-1

\begin{acks}
This work was partially supported by the China Scholarship Council (202308440220). It is a result of the Information Retrieval 2 course at the University of Amsterdam. We would like to express our gratitude to Mohammad Aliannejadi and Panagiotis Eustratiadis for their organization and guidance throughout the course.
\end{acks}

\newpage
\bibliographystyle{ACM-Reference-Format}
\balance
\bibliography{main}


\begin{thebibliography}{55}


\ifx \showCODEN    \undefined \def \showCODEN     #1{\unskip}     \fi
\ifx \showDOI      \undefined \def \showDOI       #1{#1}\fi
\ifx \showISBNx    \undefined \def \showISBNx     #1{\unskip}     \fi
\ifx \showISBNxiii \undefined \def \showISBNxiii  #1{\unskip}     \fi
\ifx \showISSN     \undefined \def \showISSN      #1{\unskip}     \fi
\ifx \showLCCN     \undefined \def \showLCCN      #1{\unskip}     \fi
\ifx \shownote     \undefined \def \shownote      #1{#1}          \fi
\ifx \showarticletitle \undefined \def \showarticletitle #1{#1}   \fi
\ifx \showURL      \undefined \def \showURL       {\relax}        \fi
\providecommand\bibfield[2]{#2}
\providecommand\bibinfo[2]{#2}
\providecommand\natexlab[1]{#1}
\providecommand\showeprint[2][]{arXiv:#2}

\bibitem[Abdalla et~al\mbox{.}(2020)]%
        {info:doi/10.2196/18055}
\bibfield{author}{\bibinfo{person}{Mohamed Abdalla}, \bibinfo{person}{Moustafa Abdalla}, \bibinfo{person}{Graeme Hirst}, {and} \bibinfo{person}{Frank Rudzicz}.} \bibinfo{year}{2020}\natexlab{}.
\newblock \showarticletitle{Exploring the Privacy-Preserving Properties of Word Embeddings: Algorithmic Validation Study}.
\newblock \bibinfo{journal}{\emph{J Med Internet Res}} \bibinfo{volume}{22}, \bibinfo{number}{7} (\bibinfo{date}{15 Jul} \bibinfo{year}{2020}), \bibinfo{pages}{e18055}.
\newblock
\showISSN{1438-8871}
\urldef\tempurl%
\url{https://doi.org/10.2196/18055}
\showDOI{\tempurl}


\bibitem[B(2019)]%
        {bhavik2019password}
\bibfield{author}{\bibinfo{person}{Bhavik B}.} \bibinfo{year}{2019}\natexlab{}.
\newblock \bibinfo{title}{Password Strength Classifier Dataset}.
\newblock \bibinfo{howpublished}{Kaggle}.
\newblock
\urldef\tempurl%
\url{https://www.kaggle.com/datasets/bhavikbb/password-strength-classifier-dataset}
\showURL{%
\tempurl}
\newblock
\shownote{Accessed: 2025-05-08}.


\bibitem[Boteva et~al\mbox{.}(2016)]%
        {BotevaGSR16_nfcorpus}
\bibfield{author}{\bibinfo{person}{Vera Boteva}, \bibinfo{person}{Demian~Gholipour Ghalandari}, \bibinfo{person}{Artem Sokolov}, {and} \bibinfo{person}{Stefan Riezler}.} \bibinfo{year}{2016}\natexlab{}.
\newblock \showarticletitle{A Full-Text Learning to Rank Dataset for Medical Information Retrieval}. In \bibinfo{booktitle}{\emph{Advances in Information Retrieval - 38th European Conference on {IR} Research, {ECIR} 2016, Padua, Italy, March 20-23, 2016. Proceedings}} \emph{(\bibinfo{series}{Lecture Notes in Computer Science}, Vol.~\bibinfo{volume}{9626})}, \bibfield{editor}{\bibinfo{person}{Nicola Ferro}, \bibinfo{person}{Fabio Crestani}, \bibinfo{person}{Marie{-}Francine Moens}, \bibinfo{person}{Josiane Mothe}, \bibinfo{person}{Fabrizio Silvestri}, \bibinfo{person}{Giorgio Maria~Di Nunzio}, \bibinfo{person}{Claudia Hauff}, {and} \bibinfo{person}{Gianmaria Silvello}} (Eds.). \bibinfo{publisher}{Springer}, \bibinfo{pages}{716--722}.
\newblock
\urldef\tempurl%
\url{https://doi.org/10.1007/978-3-319-30671-1\_58}
\showDOI{\tempurl}


\bibitem[Chen et~al\mbox{.}(2024)]%
        {chen2024textembeddinginversionsecurity}
\bibfield{author}{\bibinfo{person}{Yiyi Chen}, \bibinfo{person}{Heather~C. Lent}, {and} \bibinfo{person}{Johannes Bjerva}.} \bibinfo{year}{2024}\natexlab{}.
\newblock \showarticletitle{Text Embedding Inversion Security for Multilingual Language Models}. In \bibinfo{booktitle}{\emph{Proceedings of the 62nd Annual Meeting of the Association for Computational Linguistics (Volume 1: Long Papers), {ACL} 2024, Bangkok, Thailand, August 11-16, 2024}}, \bibfield{editor}{\bibinfo{person}{Lun{-}Wei Ku}, \bibinfo{person}{Andre Martins}, {and} \bibinfo{person}{Vivek Srikumar}} (Eds.). \bibinfo{publisher}{Association for Computational Linguistics}, \bibinfo{pages}{7808--7827}.
\newblock
\urldef\tempurl%
\url{https://doi.org/10.18653/V1/2024.ACL-LONG.422}
\showDOI{\tempurl}


\bibitem[Dettmers et~al\mbox{.}(2022)]%
        {NEURIPS2022_8_bit_Quantization}
\bibfield{author}{\bibinfo{person}{Tim Dettmers}, \bibinfo{person}{Mike Lewis}, \bibinfo{person}{Younes Belkada}, {and} \bibinfo{person}{Luke Zettlemoyer}.} \bibinfo{year}{2022}\natexlab{}.
\newblock \showarticletitle{GPT3.int8(): 8-bit Matrix Multiplication for Transformers at Scale}. In \bibinfo{booktitle}{\emph{Advances in Neural Information Processing Systems}}, \bibfield{editor}{\bibinfo{person}{S.~Koyejo}, \bibinfo{person}{S.~Mohamed}, \bibinfo{person}{A.~Agarwal}, \bibinfo{person}{D.~Belgrave}, \bibinfo{person}{K.~Cho}, {and} \bibinfo{person}{A.~Oh}} (Eds.), Vol.~\bibinfo{volume}{35}. \bibinfo{publisher}{Curran Associates, Inc.}, \bibinfo{pages}{30318--30332}.
\newblock
\urldef\tempurl%
\url{https://proceedings.neurips.cc/paper_files/paper/2022/file/c3ba4962c05c49636d4c6206a97e9c8a-Paper-Conference.pdf}
\showURL{%
\tempurl}


\bibitem[Devlin et~al\mbox{.}(2019)]%
        {naaclDevlinCLT19_BERT}
\bibfield{author}{\bibinfo{person}{Jacob Devlin}, \bibinfo{person}{Ming{-}Wei Chang}, \bibinfo{person}{Kenton Lee}, {and} \bibinfo{person}{Kristina Toutanova}.} \bibinfo{year}{2019}\natexlab{}.
\newblock \showarticletitle{{BERT:} Pre-training of Deep Bidirectional Transformers for Language Understanding}. In \bibinfo{booktitle}{\emph{Proceedings of the 2019 Conference of the North American Chapter of the Association for Computational Linguistics: Human Language Technologies, {NAACL-HLT} 2019, Minneapolis, MN, USA, June 2-7, 2019, Volume 1 (Long and Short Papers)}}, \bibfield{editor}{\bibinfo{person}{Jill Burstein}, \bibinfo{person}{Christy Doran}, {and} \bibinfo{person}{Thamar Solorio}} (Eds.). \bibinfo{publisher}{Association for Computational Linguistics}, \bibinfo{pages}{4171--4186}.
\newblock
\urldef\tempurl%
\url{https://doi.org/10.18653/V1/N19-1423}
\showDOI{\tempurl}


\bibitem[Diggelmann et~al\mbox{.}(2020)]%
        {abs-2012-00614_climate_fever}
\bibfield{author}{\bibinfo{person}{Thomas Diggelmann}, \bibinfo{person}{Jordan~L. Boyd{-}Graber}, \bibinfo{person}{Jannis Bulian}, \bibinfo{person}{Massimiliano Ciaramita}, {and} \bibinfo{person}{Markus Leippold}.} \bibinfo{year}{2020}\natexlab{}.
\newblock \showarticletitle{{CLIMATE-FEVER:} {A} Dataset for Verification of Real-World Climate Claims}.
\newblock \bibinfo{journal}{\emph{CoRR}}  \bibinfo{volume}{abs/2012.00614} (\bibinfo{year}{2020}).
\newblock
\showeprint[arXiv]{2012.00614}
\urldef\tempurl%
\url{https://arxiv.org/abs/2012.00614}
\showURL{%
\tempurl}


\bibitem[Dosovitskiy and Brox(2016)]%
        {cvpr_DosovitskiyB16}
\bibfield{author}{\bibinfo{person}{Alexey Dosovitskiy} {and} \bibinfo{person}{Thomas Brox}.} \bibinfo{year}{2016}\natexlab{}.
\newblock \showarticletitle{Inverting Visual Representations with Convolutional Networks}. In \bibinfo{booktitle}{\emph{2016 {IEEE} Conference on Computer Vision and Pattern Recognition, {CVPR} 2016, Las Vegas, NV, USA, June 27-30, 2016}}. \bibinfo{publisher}{{IEEE} Computer Society}, \bibinfo{pages}{4829--4837}.
\newblock
\urldef\tempurl%
\url{https://doi.org/10.1109/CVPR.2016.522}
\showDOI{\tempurl}


\bibitem[Douze et~al\mbox{.}(2024)]%
        {douze2024faiss}
\bibfield{author}{\bibinfo{person}{Matthijs Douze}, \bibinfo{person}{Alexandr Guzhva}, \bibinfo{person}{Chengqi Deng}, \bibinfo{person}{Jeff Johnson}, \bibinfo{person}{Gergely Szilvasy}, \bibinfo{person}{Pierre-Emmanuel Mazaré}, \bibinfo{person}{Maria Lomeli}, \bibinfo{person}{Lucas Hosseini}, {and} \bibinfo{person}{Hervé Jégou}.} \bibinfo{year}{2024}\natexlab{}.
\newblock \showarticletitle{The Faiss library}.
\newblock  (\bibinfo{year}{2024}).
\newblock
\showeprint[arxiv]{2401.08281}~[cs.LG]


\bibitem[Iyer et~al\mbox{.}(2017)]%
        {quora2017dataset}
\bibfield{author}{\bibinfo{person}{Shankar Iyer}, \bibinfo{person}{Nikhil Dandekar}, {and} \bibinfo{person}{Kornél Csernai}.} \bibinfo{year}{2017}\natexlab{}.
\newblock \bibinfo{title}{First Quora Dataset Release: Question Pairs}.
\newblock
\newblock
\urldef\tempurl%
\url{https://quoradata.quora.com/First-Quora-Dataset-Release-Question-Pairs}
\showURL{%
\tempurl}


\bibitem[Johnson et~al\mbox{.}(2016)]%
        {johnson2016mimic}
\bibfield{author}{\bibinfo{person}{Alistair~EW Johnson}, \bibinfo{person}{Tom~J Pollard}, \bibinfo{person}{Lu Shen}, \bibinfo{person}{Li-wei~H Lehman}, \bibinfo{person}{Mengling Feng}, \bibinfo{person}{Mohammad Ghassemi}, \bibinfo{person}{Benjamin Moody}, \bibinfo{person}{Peter Szolovits}, \bibinfo{person}{Leo Anthony~Celi}, {and} \bibinfo{person}{Roger~G Mark}.} \bibinfo{year}{2016}\natexlab{}.
\newblock \showarticletitle{MIMIC-III, a freely accessible critical care database}.
\newblock \bibinfo{journal}{\emph{Scientific data}} \bibinfo{volume}{3}, \bibinfo{number}{1} (\bibinfo{year}{2016}), \bibinfo{pages}{1--9}.
\newblock
\urldef\tempurl%
\url{https://doi.org/10.1038/sdata.2016.35}
\showDOI{\tempurl}


\bibitem[Karpukhin et~al\mbox{.}(2020)]%
        {KarpukhinOMLWEC20_DPR}
\bibfield{author}{\bibinfo{person}{Vladimir Karpukhin}, \bibinfo{person}{Barlas Oguz}, \bibinfo{person}{Sewon Min}, \bibinfo{person}{Patrick S.~H. Lewis}, \bibinfo{person}{Ledell Wu}, \bibinfo{person}{Sergey Edunov}, \bibinfo{person}{Danqi Chen}, {and} \bibinfo{person}{Wen{-}tau Yih}.} \bibinfo{year}{2020}\natexlab{}.
\newblock \showarticletitle{Dense Passage Retrieval for Open-Domain Question Answering}. In \bibinfo{booktitle}{\emph{Proceedings of the 2020 Conference on Empirical Methods in Natural Language Processing, {EMNLP} 2020}}. \bibinfo{publisher}{Association for Computational Linguistics}, \bibinfo{pages}{6769--6781}.
\newblock
\urldef\tempurl%
\url{https://doi.org/10.18653/V1/2020.EMNLP-MAIN.550}
\showDOI{\tempurl}


\bibitem[Kastaniotis(2022)]%
        {weight_quantization}
\bibfield{author}{\bibinfo{person}{Dimitris Kastaniotis}.} \bibinfo{year}{2022}\natexlab{}.
\newblock \bibinfo{title}{Introduction to Weight Quantization}.
\newblock \bibinfo{howpublished}{Towards Data Science}.
\newblock
\urldef\tempurl%
\url{https://towardsdatascience.com/introduction-to-weight-quantization-2494701b9c0c/}
\showURL{%
\tempurl}
\newblock
\shownote{Accessed: \today}.


\bibitem[Khattab and Zaharia(2020)]%
        {KhattabZ20_ColBERT}
\bibfield{author}{\bibinfo{person}{Omar Khattab} {and} \bibinfo{person}{Matei Zaharia}.} \bibinfo{year}{2020}\natexlab{}.
\newblock \showarticletitle{ColBERT: Efficient and Effective Passage Search via Contextualized Late Interaction over {BERT}}. In \bibinfo{booktitle}{\emph{Proceedings of the 43rd International {ACM} {SIGIR} conference on research and development in Information Retrieval, {SIGIR} 2020}}. \bibinfo{publisher}{{ACM}}, \bibinfo{pages}{39--48}.
\newblock
\urldef\tempurl%
\url{https://doi.org/10.1145/3397271.3401075}
\showDOI{\tempurl}


\bibitem[Kiros et~al\mbox{.}(2015)]%
        {nips_KirosZSZUTF15}
\bibfield{author}{\bibinfo{person}{Ryan Kiros}, \bibinfo{person}{Yukun Zhu}, \bibinfo{person}{Ruslan Salakhutdinov}, \bibinfo{person}{Richard~S. Zemel}, \bibinfo{person}{Raquel Urtasun}, \bibinfo{person}{Antonio Torralba}, {and} \bibinfo{person}{Sanja Fidler}.} \bibinfo{year}{2015}\natexlab{}.
\newblock \showarticletitle{Skip-Thought Vectors}. In \bibinfo{booktitle}{\emph{Advances in Neural Information Processing Systems 28: Annual Conference on Neural Information Processing Systems 2015, December 7-12, 2015, Montreal, Quebec, Canada}}, \bibfield{editor}{\bibinfo{person}{Corinna Cortes}, \bibinfo{person}{Neil~D. Lawrence}, \bibinfo{person}{Daniel~D. Lee}, \bibinfo{person}{Masashi Sugiyama}, {and} \bibinfo{person}{Roman Garnett}} (Eds.). \bibinfo{pages}{3294--3302}.
\newblock
\urldef\tempurl%
\url{https://proceedings.neurips.cc/paper/2015/hash/f442d33fa06832082290ad8544a8da27-Abstract.html}
\showURL{%
\tempurl}


\bibitem[Kwiatkowski et~al\mbox{.}(2019)]%
        {KwiatkowskiPRCP19_NQ}
\bibfield{author}{\bibinfo{person}{Tom Kwiatkowski}, \bibinfo{person}{Jennimaria Palomaki}, \bibinfo{person}{Olivia Redfield}, \bibinfo{person}{Michael Collins}, \bibinfo{person}{Ankur~P. Parikh}, \bibinfo{person}{Chris Alberti}, \bibinfo{person}{Danielle Epstein}, \bibinfo{person}{Illia Polosukhin}, \bibinfo{person}{Jacob Devlin}, \bibinfo{person}{Kenton Lee}, \bibinfo{person}{Kristina Toutanova}, \bibinfo{person}{Llion Jones}, \bibinfo{person}{Matthew Kelcey}, \bibinfo{person}{Ming{-}Wei Chang}, \bibinfo{person}{Andrew~M. Dai}, \bibinfo{person}{Jakob Uszkoreit}, \bibinfo{person}{Quoc Le}, {and} \bibinfo{person}{Slav Petrov}.} \bibinfo{year}{2019}\natexlab{}.
\newblock \showarticletitle{Natural Questions: a Benchmark for Question Answering Research}.
\newblock \bibinfo{journal}{\emph{Trans. Assoc. Comput. Linguistics}}  \bibinfo{volume}{7} (\bibinfo{year}{2019}), \bibinfo{pages}{452--466}.
\newblock
\urldef\tempurl%
\url{https://doi.org/10.1162/TACL\_A\_00276}
\showDOI{\tempurl}


\bibitem[Le and Mikolov(2014)]%
        {icml_LeM14}
\bibfield{author}{\bibinfo{person}{Quoc~V. Le} {and} \bibinfo{person}{Tom{\'{a}}s Mikolov}.} \bibinfo{year}{2014}\natexlab{}.
\newblock \showarticletitle{Distributed Representations of Sentences and Documents}. In \bibinfo{booktitle}{\emph{Proceedings of the 31th International Conference on Machine Learning, {ICML} 2014, Beijing, China, 21-26 June 2014}} \emph{(\bibinfo{series}{{JMLR} Workshop and Conference Proceedings}, Vol.~\bibinfo{volume}{32})}. \bibinfo{publisher}{JMLR.org}, \bibinfo{pages}{1188--1196}.
\newblock
\urldef\tempurl%
\url{http://proceedings.mlr.press/v32/le14.html}
\showURL{%
\tempurl}


\bibitem[Lehman et~al\mbox{.}(2021)]%
        {lehman2021doesbertpretrainedclinical}
\bibfield{author}{\bibinfo{person}{Eric Lehman}, \bibinfo{person}{Sarthak Jain}, \bibinfo{person}{Karl Pichotta}, \bibinfo{person}{Yoav Goldberg}, {and} \bibinfo{person}{Byron~C. Wallace}.} \bibinfo{year}{2021}\natexlab{}.
\newblock \showarticletitle{Does {BERT} Pretrained on Clinical Notes Reveal Sensitive Data?}. In \bibinfo{booktitle}{\emph{Proceedings of the 2021 Conference of the North American Chapter of the Association for Computational Linguistics: Human Language Technologies, {NAACL-HLT} 2021, Online, June 6-11, 2021}}, \bibfield{editor}{\bibinfo{person}{Kristina Toutanova}, \bibinfo{person}{Anna Rumshisky}, \bibinfo{person}{Luke Zettlemoyer}, \bibinfo{person}{Dilek Hakkani{-}T{\"{u}}r}, \bibinfo{person}{Iz~Beltagy}, \bibinfo{person}{Steven Bethard}, \bibinfo{person}{Ryan Cotterell}, \bibinfo{person}{Tanmoy Chakraborty}, {and} \bibinfo{person}{Yichao Zhou}} (Eds.). \bibinfo{publisher}{Association for Computational Linguistics}, \bibinfo{pages}{946--959}.
\newblock
\urldef\tempurl%
\url{https://doi.org/10.18653/V1/2021.NAACL-MAIN.73}
\showDOI{\tempurl}


\bibitem[Lei et~al\mbox{.}(2025)]%
        {Yibing_LENS}
\bibfield{author}{\bibinfo{person}{Yibin Lei}, \bibinfo{person}{Tao Shen}, \bibinfo{person}{Yu Cao}, {and} \bibinfo{person}{Andrew Yates}.} \bibinfo{year}{2025}\natexlab{}.
\newblock \showarticletitle{Enhancing Lexicon-Based Text Embeddings with Large Language Models}.
\newblock \bibinfo{journal}{\emph{CoRR}}  \bibinfo{volume}{abs/2501.09749} (\bibinfo{year}{2025}).
\newblock
\urldef\tempurl%
\url{https://doi.org/10.48550/ARXIV.2501.09749}
\showDOI{\tempurl}
\showeprint[arXiv]{2501.09749}


\bibitem[Lei et~al\mbox{.}(2024)]%
        {Yibin_MetaEOL}
\bibfield{author}{\bibinfo{person}{Yibin Lei}, \bibinfo{person}{Di Wu}, \bibinfo{person}{Tianyi Zhou}, \bibinfo{person}{Tao Shen}, \bibinfo{person}{Yu Cao}, \bibinfo{person}{Chongyang Tao}, {and} \bibinfo{person}{Andrew Yates}.} \bibinfo{year}{2024}\natexlab{}.
\newblock \showarticletitle{Meta-Task Prompting Elicits Embeddings from Large Language Models}. In \bibinfo{booktitle}{\emph{Proceedings of the 62nd Annual Meeting of the Association for Computational Linguistics (Volume 1: Long Papers), {ACL} 2024, Bangkok, Thailand, August 11-16, 2024}}, \bibfield{editor}{\bibinfo{person}{Lun{-}Wei Ku}, \bibinfo{person}{Andre Martins}, {and} \bibinfo{person}{Vivek Srikumar}} (Eds.). \bibinfo{publisher}{Association for Computational Linguistics}, \bibinfo{pages}{10141--10157}.
\newblock
\urldef\tempurl%
\url{https://doi.org/10.18653/V1/2024.ACL-LONG.546}
\showDOI{\tempurl}


\bibitem[Li et~al\mbox{.}(2023)]%
        {li2023sentenceembeddingleaksinformation}
\bibfield{author}{\bibinfo{person}{Haoran Li}, \bibinfo{person}{Mingshi Xu}, {and} \bibinfo{person}{Yangqiu Song}.} \bibinfo{year}{2023}\natexlab{}.
\newblock \showarticletitle{Sentence Embedding Leaks More Information than You Expect: Generative Embedding Inversion Attack to Recover the Whole Sentence}. In \bibinfo{booktitle}{\emph{Findings of the Association for Computational Linguistics: {ACL} 2023, Toronto, Canada, July 9-14, 2023}}, \bibfield{editor}{\bibinfo{person}{Anna Rogers}, \bibinfo{person}{Jordan~L. Boyd{-}Graber}, {and} \bibinfo{person}{Naoaki Okazaki}} (Eds.). \bibinfo{publisher}{Association for Computational Linguistics}, \bibinfo{pages}{14022--14040}.
\newblock
\urldef\tempurl%
\url{https://doi.org/10.18653/V1/2023.FINDINGS-ACL.881}
\showDOI{\tempurl}


\bibitem[Li et~al\mbox{.}(2025a)]%
        {li2025reproducinghotflip}
\bibfield{author}{\bibinfo{person}{Yongkang Li}, \bibinfo{person}{Panagiotis Eustratiadis}, {and} \bibinfo{person}{Evangelos Kanoulas}.} \bibinfo{year}{2025}\natexlab{a}.
\newblock \showarticletitle{Reproducing HotFlip for Corpus Poisoning Attacks in Dense Retrieval}. In \bibinfo{booktitle}{\emph{Advances in Information Retrieval - 47th European Conference on Information Retrieval, {ECIR} 2025, Lucca, Italy, April 6-10, 2025, Proceedings, Part {IV}}} \emph{(\bibinfo{series}{Lecture Notes in Computer Science}, Vol.~\bibinfo{volume}{15575})}. \bibinfo{publisher}{Springer}, \bibinfo{pages}{95--111}.
\newblock
\urldef\tempurl%
\url{https://doi.org/10.1007/978-3-031-88717-8\_8}
\showDOI{\tempurl}


\bibitem[Li et~al\mbox{.}(2025b)]%
        {li2025unsupervisedcorpuspoisoningattacks}
\bibfield{author}{\bibinfo{person}{Yongkang Li}, \bibinfo{person}{Panagiotis Eustratiadis}, \bibinfo{person}{Simon Lupart}, {and} \bibinfo{person}{Evangelos Kanoulas}.} \bibinfo{year}{2025}\natexlab{b}.
\newblock \showarticletitle{Unsupervised Corpus Poisoning Attacks in Continuous Space for Dense Retrieval}. In \bibinfo{booktitle}{\emph{Proceedings of the 48th International ACM SIGIR Conference on Research and Development in Information Retrieval, {SIGIR} 2025}}.
\newblock
\urldef\tempurl%
\url{https://doi.org/10.1145/3726302.3730110}
\showDOI{\tempurl}


\bibitem[Ma et~al\mbox{.}(2024)]%
        {MaWYWL24_RankLLaMA}
\bibfield{author}{\bibinfo{person}{Xueguang Ma}, \bibinfo{person}{Liang Wang}, \bibinfo{person}{Nan Yang}, \bibinfo{person}{Furu Wei}, {and} \bibinfo{person}{Jimmy Lin}.} \bibinfo{year}{2024}\natexlab{}.
\newblock \showarticletitle{Fine-Tuning LLaMA for Multi-Stage Text Retrieval}. In \bibinfo{booktitle}{\emph{Proceedings of the 47th International {ACM} {SIGIR} Conference on Research and Development in Information Retrieval, {SIGIR} 2024, Washington DC, USA, July 14-18, 2024}}, \bibfield{editor}{\bibinfo{person}{Grace~Hui Yang}, \bibinfo{person}{Hongning Wang}, \bibinfo{person}{Sam Han}, \bibinfo{person}{Claudia Hauff}, \bibinfo{person}{Guido Zuccon}, {and} \bibinfo{person}{Yi~Zhang}} (Eds.). \bibinfo{publisher}{{ACM}}, \bibinfo{pages}{2421--2425}.
\newblock
\urldef\tempurl%
\url{https://doi.org/10.1145/3626772.3657951}
\showDOI{\tempurl}


\bibitem[Mahendran and Vedaldi(2015)]%
        {cvpr_MahendranV15}
\bibfield{author}{\bibinfo{person}{Aravindh Mahendran} {and} \bibinfo{person}{Andrea Vedaldi}.} \bibinfo{year}{2015}\natexlab{}.
\newblock \showarticletitle{Understanding deep image representations by inverting them}. In \bibinfo{booktitle}{\emph{{IEEE} Conference on Computer Vision and Pattern Recognition, {CVPR} 2015, Boston, MA, USA, June 7-12, 2015}}. \bibinfo{publisher}{{IEEE} Computer Society}, \bibinfo{pages}{5188--5196}.
\newblock
\urldef\tempurl%
\url{https://doi.org/10.1109/CVPR.2015.7299155}
\showDOI{\tempurl}


\bibitem[Maia et~al\mbox{.}(2018)]%
        {MaiaHFDMZB18_fiqa}
\bibfield{author}{\bibinfo{person}{Macedo Maia}, \bibinfo{person}{Siegfried Handschuh}, \bibinfo{person}{Andr{\'{e}} Freitas}, \bibinfo{person}{Brian Davis}, \bibinfo{person}{Ross McDermott}, \bibinfo{person}{Manel Zarrouk}, {and} \bibinfo{person}{Alexandra Balahur}.} \bibinfo{year}{2018}\natexlab{}.
\newblock \showarticletitle{WWW'18 Open Challenge: Financial Opinion Mining and Question Answering}. In \bibinfo{booktitle}{\emph{Companion of the The Web Conference 2018 on The Web Conference 2018, {WWW} 2018, Lyon , France, April 23-27, 2018}}, \bibfield{editor}{\bibinfo{person}{Pierre{-}Antoine Champin}, \bibinfo{person}{Fabien Gandon}, \bibinfo{person}{Mounia Lalmas}, {and} \bibinfo{person}{Panagiotis~G. Ipeirotis}} (Eds.). \bibinfo{publisher}{{ACM}}, \bibinfo{pages}{1941--1942}.
\newblock
\urldef\tempurl%
\url{https://doi.org/10.1145/3184558.3192301}
\showURL{%
\tempurl}


\bibitem[Mathur et~al\mbox{.}(2019)]%
        {acl_MathurBC19}
\bibfield{author}{\bibinfo{person}{Nitika Mathur}, \bibinfo{person}{Timothy Baldwin}, {and} \bibinfo{person}{Trevor Cohn}.} \bibinfo{year}{2019}\natexlab{}.
\newblock \showarticletitle{Putting Evaluation in Context: Contextual Embeddings Improve Machine Translation Evaluation}. In \bibinfo{booktitle}{\emph{Proceedings of the 57th Conference of the Association for Computational Linguistics, {ACL} 2019, Florence, Italy, July 28- August 2, 2019, Volume 1: Long Papers}}, \bibfield{editor}{\bibinfo{person}{Anna Korhonen}, \bibinfo{person}{David~R. Traum}, {and} \bibinfo{person}{Llu{\'{\i}}s M{\`{a}}rquez}} (Eds.). \bibinfo{publisher}{Association for Computational Linguistics}, \bibinfo{pages}{2799--2808}.
\newblock
\urldef\tempurl%
\url{https://doi.org/10.18653/V1/P19-1269}
\showDOI{\tempurl}


\bibitem[Mi et~al\mbox{.}(2016)]%
        {emnlp_MiSWI16}
\bibfield{author}{\bibinfo{person}{Haitao Mi}, \bibinfo{person}{Baskaran Sankaran}, \bibinfo{person}{Zhiguo Wang}, {and} \bibinfo{person}{Abe Ittycheriah}.} \bibinfo{year}{2016}\natexlab{}.
\newblock \showarticletitle{Coverage Embedding Models for Neural Machine Translation}. In \bibinfo{booktitle}{\emph{Proceedings of the 2016 Conference on Empirical Methods in Natural Language Processing, {EMNLP} 2016, Austin, Texas, USA, November 1-4, 2016}}, \bibfield{editor}{\bibinfo{person}{Jian Su}, \bibinfo{person}{Xavier Carreras}, {and} \bibinfo{person}{Kevin Duh}} (Eds.). \bibinfo{publisher}{The Association for Computational Linguistics}, \bibinfo{pages}{955--960}.
\newblock
\urldef\tempurl%
\url{https://doi.org/10.18653/V1/D16-1096}
\showDOI{\tempurl}


\bibitem[Mikolov et~al\mbox{.}(2013)]%
        {Word2Vec}
\bibfield{author}{\bibinfo{person}{Tom{\'{a}}s Mikolov}, \bibinfo{person}{Kai Chen}, \bibinfo{person}{Greg Corrado}, {and} \bibinfo{person}{Jeffrey Dean}.} \bibinfo{year}{2013}\natexlab{}.
\newblock \showarticletitle{Efficient Estimation of Word Representations in Vector Space}. In \bibinfo{booktitle}{\emph{1st International Conference on Learning Representations, {ICLR} 2013, Scottsdale, Arizona, USA, May 2-4, 2013, Workshop Track Proceedings}}, \bibfield{editor}{\bibinfo{person}{Yoshua Bengio} {and} \bibinfo{person}{Yann LeCun}} (Eds.).
\newblock
\urldef\tempurl%
\url{http://arxiv.org/abs/1301.3781}
\showURL{%
\tempurl}


\bibitem[Morris et~al\mbox{.}(2023)]%
        {morris2023text}
\bibfield{author}{\bibinfo{person}{John~X. Morris}, \bibinfo{person}{Volodymyr Kuleshov}, \bibinfo{person}{Vitaly Shmatikov}, {and} \bibinfo{person}{Alexander~M. Rush}.} \bibinfo{year}{2023}\natexlab{}.
\newblock \showarticletitle{Text Embeddings Reveal (Almost) As Much As Text}. In \bibinfo{booktitle}{\emph{Proceedings of the 2023 Conference on Empirical Methods in Natural Language Processing, {EMNLP} 2023, Singapore, December 6-10, 2023}}, \bibfield{editor}{\bibinfo{person}{Houda Bouamor}, \bibinfo{person}{Juan Pino}, {and} \bibinfo{person}{Kalika Bali}} (Eds.). \bibinfo{publisher}{Association for Computational Linguistics}, \bibinfo{pages}{12448--12460}.
\newblock
\urldef\tempurl%
\url{https://doi.org/10.18653/V1/2023.EMNLP-MAIN.765}
\showDOI{\tempurl}


\bibitem[Nguyen et~al\mbox{.}(2016)]%
        {msmarco}
\bibfield{author}{\bibinfo{person}{Tri Nguyen}, \bibinfo{person}{Mir Rosenberg}, \bibinfo{person}{Xia Song}, \bibinfo{person}{Jianfeng Gao}, \bibinfo{person}{Saurabh Tiwary}, \bibinfo{person}{Rangan Majumder}, {and} \bibinfo{person}{Li Deng}.} \bibinfo{year}{2016}\natexlab{}.
\newblock \showarticletitle{{MS} {MARCO:} {A} Human Generated MAchine Reading COmprehension Dataset}.
\newblock \bibinfo{journal}{\emph{CoRR}}  \bibinfo{volume}{abs/1611.09268} (\bibinfo{year}{2016}).
\newblock
\showeprint[arxiv]{1611.09268}
\urldef\tempurl%
\url{http://arxiv.org/abs/1611.09268}
\showURL{%
\tempurl}


\bibitem[Ni et~al\mbox{.}(2022)]%
        {ni2021large}
\bibfield{author}{\bibinfo{person}{Jianmo Ni}, \bibinfo{person}{Chen Qu}, \bibinfo{person}{Jing Lu}, \bibinfo{person}{Zhuyun Dai}, \bibinfo{person}{Gustavo~Hern{\'{a}}ndez {\'{A}}brego}, \bibinfo{person}{Ji Ma}, \bibinfo{person}{Vincent~Y. Zhao}, \bibinfo{person}{Yi Luan}, \bibinfo{person}{Keith~B. Hall}, \bibinfo{person}{Ming{-}Wei Chang}, {and} \bibinfo{person}{Yinfei Yang}.} \bibinfo{year}{2022}\natexlab{}.
\newblock \showarticletitle{Large Dual Encoders Are Generalizable Retrievers}. In \bibinfo{booktitle}{\emph{Proceedings of the 2022 Conference on Empirical Methods in Natural Language Processing, {EMNLP} 2022, Abu Dhabi, United Arab Emirates, December 7-11, 2022}}, \bibfield{editor}{\bibinfo{person}{Yoav Goldberg}, \bibinfo{person}{Zornitsa Kozareva}, {and} \bibinfo{person}{Yue Zhang}} (Eds.). \bibinfo{publisher}{Association for Computational Linguistics}, \bibinfo{pages}{9844--9855}.
\newblock
\urldef\tempurl%
\url{https://doi.org/10.18653/V1/2022.EMNLP-MAIN.669}
\showDOI{\tempurl}


\bibitem[Okura et~al\mbox{.}(2017)]%
        {kdd_OkuraTOT17}
\bibfield{author}{\bibinfo{person}{Shumpei Okura}, \bibinfo{person}{Yukihiro Tagami}, \bibinfo{person}{Shingo Ono}, {and} \bibinfo{person}{Akira Tajima}.} \bibinfo{year}{2017}\natexlab{}.
\newblock \showarticletitle{Embedding-based News Recommendation for Millions of Users}. In \bibinfo{booktitle}{\emph{Proceedings of the 23rd {ACM} {SIGKDD} International Conference on Knowledge Discovery and Data Mining, Halifax, NS, Canada, August 13 - 17, 2017}}. \bibinfo{publisher}{{ACM}}, \bibinfo{pages}{1933--1942}.
\newblock
\urldef\tempurl%
\url{https://doi.org/10.1145/3097983.3098108}
\showDOI{\tempurl}


\bibitem[{OpenAI}(2022)]%
        {adaopen}
\bibfield{author}{\bibinfo{person}{{OpenAI}}.} \bibinfo{year}{2022}\natexlab{}.
\newblock \bibinfo{title}{New and improved embedding model}.
\newblock \bibinfo{howpublished}{\url{https://openai.com/index/new-and-improved-embedding-model/}}.
\newblock
\newblock
\shownote{Accessed: 2025-05-06}.


\bibitem[Papineni et~al\mbox{.}(2002)]%
        {PapineniRWZ02_BLEU}
\bibfield{author}{\bibinfo{person}{Kishore Papineni}, \bibinfo{person}{Salim Roukos}, \bibinfo{person}{Todd Ward}, {and} \bibinfo{person}{Wei{-}Jing Zhu}.} \bibinfo{year}{2002}\natexlab{}.
\newblock \showarticletitle{Bleu: a Method for Automatic Evaluation of Machine Translation}. In \bibinfo{booktitle}{\emph{Proceedings of the 40th Annual Meeting of the Association for Computational Linguistics, July 6-12, 2002, Philadelphia, PA, {USA}}}. \bibinfo{publisher}{{ACL}}, \bibinfo{pages}{311--318}.
\newblock
\urldef\tempurl%
\url{https://doi.org/10.3115/1073083.1073135}
\showDOI{\tempurl}


\bibitem[Parikh et~al\mbox{.}(2022)]%
        {parikh2022canaryextractionnaturallanguage}
\bibfield{author}{\bibinfo{person}{Rahil Parikh}, \bibinfo{person}{Christophe Dupuy}, {and} \bibinfo{person}{Rahul Gupta}.} \bibinfo{year}{2022}\natexlab{}.
\newblock \showarticletitle{Canary Extraction in Natural Language Understanding Models}. In \bibinfo{booktitle}{\emph{Proceedings of the 60th Annual Meeting of the Association for Computational Linguistics (Volume 2: Short Papers), {ACL} 2022, Dublin, Ireland, May 22-27, 2022}}, \bibfield{editor}{\bibinfo{person}{Smaranda Muresan}, \bibinfo{person}{Preslav Nakov}, {and} \bibinfo{person}{Aline Villavicencio}} (Eds.). \bibinfo{publisher}{Association for Computational Linguistics}, \bibinfo{pages}{552--560}.
\newblock
\urldef\tempurl%
\url{https://doi.org/10.18653/V1/2022.ACL-SHORT.61}
\showDOI{\tempurl}


\bibitem[Pennington et~al\mbox{.}(2014)]%
        {emnlp_PenningtonSM14}
\bibfield{author}{\bibinfo{person}{Jeffrey Pennington}, \bibinfo{person}{Richard Socher}, {and} \bibinfo{person}{Christopher~D. Manning}.} \bibinfo{year}{2014}\natexlab{}.
\newblock \showarticletitle{Glove: Global Vectors for Word Representation}. In \bibinfo{booktitle}{\emph{Proceedings of the 2014 Conference on Empirical Methods in Natural Language Processing, {EMNLP} 2014, October 25-29, 2014, Doha, Qatar, {A} meeting of SIGDAT, a Special Interest Group of the {ACL}}}, \bibfield{editor}{\bibinfo{person}{Alessandro Moschitti}, \bibinfo{person}{Bo~Pang}, {and} \bibinfo{person}{Walter Daelemans}} (Eds.). \bibinfo{publisher}{{ACL}}, \bibinfo{pages}{1532--1543}.
\newblock
\urldef\tempurl%
\url{https://doi.org/10.3115/V1/D14-1162}
\showDOI{\tempurl}


\bibitem[{Qdrant Team}(2023)]%
        {qdrant2023}
\bibfield{author}{\bibinfo{person}{{Qdrant Team}}.} \bibinfo{year}{2023}\natexlab{}.
\newblock \bibinfo{title}{Qdrant: High-performance, massive-scale vector database and vector search engine}.
\newblock \bibinfo{howpublished}{\url{https://qdrant.tech/}}.
\newblock
\newblock
\shownote{Accessed: 2025-05-06}.


\bibitem[Raffel et~al\mbox{.}(2020)]%
        {RaffelSRLNMZLL20_t5}
\bibfield{author}{\bibinfo{person}{Colin Raffel}, \bibinfo{person}{Noam Shazeer}, \bibinfo{person}{Adam Roberts}, \bibinfo{person}{Katherine Lee}, \bibinfo{person}{Sharan Narang}, \bibinfo{person}{Michael Matena}, \bibinfo{person}{Yanqi Zhou}, \bibinfo{person}{Wei Li}, {and} \bibinfo{person}{Peter~J. Liu}.} \bibinfo{year}{2020}\natexlab{}.
\newblock \showarticletitle{Exploring the Limits of Transfer Learning with a Unified Text-to-Text Transformer}.
\newblock \bibinfo{journal}{\emph{J. Mach. Learn. Res.}}  \bibinfo{volume}{21} (\bibinfo{year}{2020}), \bibinfo{pages}{140:1--140:67}.
\newblock
\urldef\tempurl%
\url{https://jmlr.org/papers/v21/20-074.html}
\showURL{%
\tempurl}


\bibitem[Soboroff et~al\mbox{.}(2018)]%
        {SoboroffHH18_trec_news}
\bibfield{author}{\bibinfo{person}{Ian Soboroff}, \bibinfo{person}{Shudong Huang}, {and} \bibinfo{person}{Donna Harman}.} \bibinfo{year}{2018}\natexlab{}.
\newblock \showarticletitle{{TREC} 2018 News Track Overview}. In \bibinfo{booktitle}{\emph{Proceedings of the Twenty-Seventh Text REtrieval Conference, {TREC} 2018, Gaithersburg, Maryland, USA, November 14-16, 2018}} \emph{(\bibinfo{series}{{NIST} Special Publication}, Vol.~\bibinfo{volume}{500-331})}, \bibfield{editor}{\bibinfo{person}{Ellen~M. Voorhees} {and} \bibinfo{person}{Angela Ellis}} (Eds.). \bibinfo{publisher}{National Institute of Standards and Technology {(NIST)}}.
\newblock
\urldef\tempurl%
\url{https://trec.nist.gov/pubs/trec27/papers/Overview-News.pdf}
\showURL{%
\tempurl}


\bibitem[Song and Raghunathan(2020)]%
        {song2020informationleakageembeddingmodels}
\bibfield{author}{\bibinfo{person}{Congzheng Song} {and} \bibinfo{person}{Ananth Raghunathan}.} \bibinfo{year}{2020}\natexlab{}.
\newblock \showarticletitle{Information Leakage in Embedding Models}. In \bibinfo{booktitle}{\emph{{CCS} '20: 2020 {ACM} {SIGSAC} Conference on Computer and Communications Security, Virtual Event, USA, November 9-13, 2020}}, \bibfield{editor}{\bibinfo{person}{Jay Ligatti}, \bibinfo{person}{Xinming Ou}, \bibinfo{person}{Jonathan Katz}, {and} \bibinfo{person}{Giovanni Vigna}} (Eds.). \bibinfo{publisher}{{ACM}}, \bibinfo{pages}{377--390}.
\newblock
\urldef\tempurl%
\url{https://doi.org/10.1145/3372297.3417270}
\showDOI{\tempurl}


\bibitem[Suarez et~al\mbox{.}(2018)]%
        {SuarezACME18_signal1m}
\bibfield{author}{\bibinfo{person}{Axel Suarez}, \bibinfo{person}{Dyaa Albakour}, \bibinfo{person}{David P.~A. Corney}, \bibinfo{person}{Miguel Martinez{-}Alvarez}, {and} \bibinfo{person}{Jos{\'{e}} Esquivel}.} \bibinfo{year}{2018}\natexlab{}.
\newblock \showarticletitle{A Data Collection for Evaluating the Retrieval of Related Tweets to News Articles}. In \bibinfo{booktitle}{\emph{Advances in Information Retrieval - 40th European Conference on {IR} Research, {ECIR} 2018, Grenoble, France, March 26-29, 2018, Proceedings}} \emph{(\bibinfo{series}{Lecture Notes in Computer Science}, Vol.~\bibinfo{volume}{10772})}, \bibfield{editor}{\bibinfo{person}{Gabriella Pasi}, \bibinfo{person}{Benjamin Piwowarski}, \bibinfo{person}{Leif Azzopardi}, {and} \bibinfo{person}{Allan Hanbury}} (Eds.). \bibinfo{publisher}{Springer}, \bibinfo{pages}{780--786}.
\newblock
\urldef\tempurl%
\url{https://doi.org/10.1007/978-3-319-76941-7\_76}
\showDOI{\tempurl}


\bibitem[Thakur et~al\mbox{.}(2021)]%
        {thakur2021beir}
\bibfield{author}{\bibinfo{person}{Nandan Thakur}, \bibinfo{person}{Nils Reimers}, \bibinfo{person}{Andreas R{\"u}ckl{\'e}}, \bibinfo{person}{Abhishek Srivastava}, {and} \bibinfo{person}{Iryna Gurevych}.} \bibinfo{year}{2021}\natexlab{}.
\newblock \showarticletitle{{BEIR}: A Heterogeneous Benchmark for Zero-shot Evaluation of Information Retrieval Models}. In \bibinfo{booktitle}{\emph{Thirty-fifth Conference on Neural Information Processing Systems Datasets and Benchmarks Track (Round 2)}}.
\newblock
\urldef\tempurl%
\url{https://openreview.net/forum?id=wCu6T5xFjeJ}
\showURL{%
\tempurl}


\bibitem[Tsatsaronis et~al\mbox{.}(2015)]%
        {TsatsaronisBMPZ15_Bioasq}
\bibfield{author}{\bibinfo{person}{George Tsatsaronis}, \bibinfo{person}{Georgios Balikas}, \bibinfo{person}{Prodromos Malakasiotis}, \bibinfo{person}{Ioannis Partalas}, \bibinfo{person}{Matthias Zschunke}, \bibinfo{person}{Michael~R. Alvers}, \bibinfo{person}{Dirk Weissenborn}, \bibinfo{person}{Anastasia Krithara}, \bibinfo{person}{Sergios Petridis}, \bibinfo{person}{Dimitris Polychronopoulos}, \bibinfo{person}{Yannis Almirantis}, \bibinfo{person}{John Pavlopoulos}, \bibinfo{person}{Nicolas Baskiotis}, \bibinfo{person}{Patrick Gallinari}, \bibinfo{person}{Thierry Arti{\`{e}}res}, \bibinfo{person}{Axel{-}Cyrille~Ngonga Ngomo}, \bibinfo{person}{Norman Heino}, \bibinfo{person}{{\'{E}}ric Gaussier}, \bibinfo{person}{Liliana Barrio{-}Alvers}, \bibinfo{person}{Michael Schroeder}, \bibinfo{person}{Ion Androutsopoulos}, {and} \bibinfo{person}{Georgios Paliouras}.} \bibinfo{year}{2015}\natexlab{}.
\newblock \showarticletitle{An overview of the {BIOASQ} large-scale biomedical semantic indexing and question answering competition}.
\newblock \bibinfo{journal}{\emph{{BMC} Bioinform.}}  \bibinfo{volume}{16} (\bibinfo{year}{2015}), \bibinfo{pages}{138:1--138:28}.
\newblock
\urldef\tempurl%
\url{https://doi.org/10.1186/S12859-015-0564-6}
\showDOI{\tempurl}


\bibitem[Veras et~al\mbox{.}(2014)]%
        {ndss_VerasCT14}
\bibfield{author}{\bibinfo{person}{Rafael Veras}, \bibinfo{person}{Christopher Collins}, {and} \bibinfo{person}{Julie Thorpe}.} \bibinfo{year}{2014}\natexlab{}.
\newblock \showarticletitle{On Semantic Patterns of Passwords and their Security Impact}. In \bibinfo{booktitle}{\emph{21st Annual Network and Distributed System Security Symposium, {NDSS} 2014, San Diego, California, USA, February 23-26, 2014}}. \bibinfo{publisher}{The Internet Society}.
\newblock
\urldef\tempurl%
\url{https://www.ndss-symposium.org/ndss2014/semantic-patterns-passwords-and-their-security-impact}
\showURL{%
\tempurl}


\bibitem[Voorhees(2003)]%
        {Voorhees03b_robust04}
\bibfield{author}{\bibinfo{person}{Ellen~M. Voorhees}.} \bibinfo{year}{2003}\natexlab{}.
\newblock \showarticletitle{Overview of the {TREC} 2003 Robust Retrieval Track}. In \bibinfo{booktitle}{\emph{Proceedings of The Twelfth Text REtrieval Conference, {TREC} 2003, Gaithersburg, Maryland, USA, November 18-21, 2003}} \emph{(\bibinfo{series}{{NIST} Special Publication}, Vol.~\bibinfo{volume}{500-255})}, \bibfield{editor}{\bibinfo{person}{Ellen~M. Voorhees} {and} \bibinfo{person}{Lori~P. Buckland}} (Eds.). \bibinfo{publisher}{National Institute of Standards and Technology {(NIST)}}, \bibinfo{pages}{69--77}.
\newblock
\urldef\tempurl%
\url{http://trec.nist.gov/pubs/trec12/papers/ROBUST.OVERVIEW.pdf}
\showURL{%
\tempurl}


\bibitem[Wachsmuth et~al\mbox{.}(2018)]%
        {WachsmuthSS18_arguana}
\bibfield{author}{\bibinfo{person}{Henning Wachsmuth}, \bibinfo{person}{Shahbaz Syed}, {and} \bibinfo{person}{Benno Stein}.} \bibinfo{year}{2018}\natexlab{}.
\newblock \showarticletitle{Retrieval of the Best Counterargument without Prior Topic Knowledge}. In \bibinfo{booktitle}{\emph{Proceedings of the 56th Annual Meeting of the Association for Computational Linguistics, {ACL} 2018, Melbourne, Australia, July 15-20, 2018, Volume 1: Long Papers}}, \bibfield{editor}{\bibinfo{person}{Iryna Gurevych} {and} \bibinfo{person}{Yusuke Miyao}} (Eds.). \bibinfo{publisher}{Association for Computational Linguistics}, \bibinfo{pages}{241--251}.
\newblock
\urldef\tempurl%
\url{https://aclanthology.org/P18-1023/}
\showURL{%
\tempurl}


\bibitem[Wadden et~al\mbox{.}(2020)]%
        {WaddenLLWZCH20_scifact}
\bibfield{author}{\bibinfo{person}{David Wadden}, \bibinfo{person}{Shanchuan Lin}, \bibinfo{person}{Kyle Lo}, \bibinfo{person}{Lucy~Lu Wang}, \bibinfo{person}{Madeleine van Zuylen}, \bibinfo{person}{Arman Cohan}, {and} \bibinfo{person}{Hannaneh Hajishirzi}.} \bibinfo{year}{2020}\natexlab{}.
\newblock \showarticletitle{Fact or Fiction: Verifying Scientific Claims}. In \bibinfo{booktitle}{\emph{Proceedings of the 2020 Conference on Empirical Methods in Natural Language Processing, {EMNLP} 2020, Online, November 16-20, 2020}}, \bibfield{editor}{\bibinfo{person}{Bonnie Webber}, \bibinfo{person}{Trevor Cohn}, \bibinfo{person}{Yulan He}, {and} \bibinfo{person}{Yang Liu}} (Eds.). \bibinfo{publisher}{Association for Computational Linguistics}, \bibinfo{pages}{7534--7550}.
\newblock
\urldef\tempurl%
\url{https://doi.org/10.18653/V1/2020.EMNLP-MAIN.609}
\showDOI{\tempurl}


\bibitem[Wang et~al\mbox{.}(2023)]%
        {WangYHJYJMW23_SimLM}
\bibfield{author}{\bibinfo{person}{Liang Wang}, \bibinfo{person}{Nan Yang}, \bibinfo{person}{Xiaolong Huang}, \bibinfo{person}{Binxing Jiao}, \bibinfo{person}{Linjun Yang}, \bibinfo{person}{Daxin Jiang}, \bibinfo{person}{Rangan Majumder}, {and} \bibinfo{person}{Furu Wei}.} \bibinfo{year}{2023}\natexlab{}.
\newblock \showarticletitle{SimLM: Pre-training with Representation Bottleneck for Dense Passage Retrieval}. In \bibinfo{booktitle}{\emph{Proceedings of the 61st Annual Meeting of the Association for Computational Linguistics (Volume 1: Long Papers), {ACL} 2023, Toronto, Canada, July 9-14, 2023}}, \bibfield{editor}{\bibinfo{person}{Anna Rogers}, \bibinfo{person}{Jordan~L. Boyd{-}Graber}, {and} \bibinfo{person}{Naoaki Okazaki}} (Eds.). \bibinfo{publisher}{Association for Computational Linguistics}, \bibinfo{pages}{2244--2258}.
\newblock
\urldef\tempurl%
\url{https://doi.org/10.18653/V1/2023.ACL-LONG.125}
\showDOI{\tempurl}


\bibitem[Wang et~al\mbox{.}(2025)]%
        {SE_PCFG}
\bibfield{author}{\bibinfo{person}{Yangde Wang}, \bibinfo{person}{Weidong Qiu}, \bibinfo{person}{Peng Tang}, \bibinfo{person}{Hao Tian}, {and} \bibinfo{person}{Shujun Li}.} \bibinfo{year}{2025}\natexlab{}.
\newblock \showarticletitle{SE{\#}PCFG: Semantically Enhanced {PCFG} for Password Analysis and Cracking}.
\newblock \bibinfo{journal}{\emph{IEEE Transactions on Dependable and Secure Computing}} (\bibinfo{year}{2025}), \bibinfo{pages}{1--14}.
\newblock
\urldef\tempurl%
\url{https://doi.org/10.1109/TDSC.2025.3547773}
\showDOI{\tempurl}


\bibitem[Wu et~al\mbox{.}(2021)]%
        {sigir_WuWQ021}
\bibfield{author}{\bibinfo{person}{Chuhan Wu}, \bibinfo{person}{Fangzhao Wu}, \bibinfo{person}{Tao Qi}, {and} \bibinfo{person}{Yongfeng Huang}.} \bibinfo{year}{2021}\natexlab{}.
\newblock \showarticletitle{Empowering News Recommendation with Pre-trained Language Models}. In \bibinfo{booktitle}{\emph{{SIGIR} '21: The 44th International {ACM} {SIGIR} Conference on Research and Development in Information Retrieval, Virtual Event, Canada, July 11-15, 2021}}, \bibfield{editor}{\bibinfo{person}{Fernando Diaz}, \bibinfo{person}{Chirag Shah}, \bibinfo{person}{Torsten Suel}, \bibinfo{person}{Pablo Castells}, \bibinfo{person}{Rosie Jones}, {and} \bibinfo{person}{Tetsuya Sakai}} (Eds.). \bibinfo{publisher}{{ACM}}, \bibinfo{pages}{1652--1656}.
\newblock
\urldef\tempurl%
\url{https://doi.org/10.1145/3404835.3463069}
\showDOI{\tempurl}


\bibitem[Xu et~al\mbox{.}(2021)]%
        {emnlp_XuDM21}
\bibfield{author}{\bibinfo{person}{Haoran Xu}, \bibinfo{person}{Benjamin~Van Durme}, {and} \bibinfo{person}{Kenton~W. Murray}.} \bibinfo{year}{2021}\natexlab{}.
\newblock \showarticletitle{BERT, mBERT, or BiBERT? {A} Study on Contextualized Embeddings for Neural Machine Translation}. In \bibinfo{booktitle}{\emph{Proceedings of the 2021 Conference on Empirical Methods in Natural Language Processing, {EMNLP} 2021, Virtual Event / Punta Cana, Dominican Republic, 7-11 November, 2021}}, \bibfield{editor}{\bibinfo{person}{Marie{-}Francine Moens}, \bibinfo{person}{Xuanjing Huang}, \bibinfo{person}{Lucia Specia}, {and} \bibinfo{person}{Scott~Wen{-}tau Yih}} (Eds.). \bibinfo{publisher}{Association for Computational Linguistics}, \bibinfo{pages}{6663--6675}.
\newblock
\urldef\tempurl%
\url{https://doi.org/10.18653/V1/2021.EMNLP-MAIN.534}
\showDOI{\tempurl}


\bibitem[Zhang et~al\mbox{.}(2016)]%
        {kdd_ZhangYLXM16}
\bibfield{author}{\bibinfo{person}{Fuzheng Zhang}, \bibinfo{person}{Nicholas~Jing Yuan}, \bibinfo{person}{Defu Lian}, \bibinfo{person}{Xing Xie}, {and} \bibinfo{person}{Wei{-}Ying Ma}.} \bibinfo{year}{2016}\natexlab{}.
\newblock \showarticletitle{Collaborative Knowledge Base Embedding for Recommender Systems}. In \bibinfo{booktitle}{\emph{Proceedings of the 22nd {ACM} {SIGKDD} International Conference on Knowledge Discovery and Data Mining, San Francisco, CA, USA, August 13-17, 2016}}, \bibfield{editor}{\bibinfo{person}{Balaji Krishnapuram}, \bibinfo{person}{Mohak Shah}, \bibinfo{person}{Alexander~J. Smola}, \bibinfo{person}{Charu~C. Aggarwal}, \bibinfo{person}{Dou Shen}, {and} \bibinfo{person}{Rajeev Rastogi}} (Eds.). \bibinfo{publisher}{{ACM}}, \bibinfo{pages}{353--362}.
\newblock
\urldef\tempurl%
\url{https://doi.org/10.1145/2939672.2939673}
\showDOI{\tempurl}


\bibitem[Zhuang et~al\mbox{.}(2024b)]%
        {zhuang2024understandingmitigatingthreatvec2text}
\bibfield{author}{\bibinfo{person}{Shengyao Zhuang}, \bibinfo{person}{Bevan Koopman}, \bibinfo{person}{Xiaoran Chu}, {and} \bibinfo{person}{Guido Zuccon}.} \bibinfo{year}{2024}\natexlab{b}.
\newblock \showarticletitle{Understanding and Mitigating the Threat of Vec2Text to Dense Retrieval Systems}. In \bibinfo{booktitle}{\emph{Proceedings of the 2024 Annual International {ACM} {SIGIR} Conference on Research and Development in Information Retrieval in the Asia Pacific Region, {SIGIR-AP} 2024, Tokyo, Japan, December 9-12, 2024}}, \bibfield{editor}{\bibinfo{person}{Tetsuya Sakai}, \bibinfo{person}{Emi Ishita}, \bibinfo{person}{Hiroaki Ohshima}, \bibinfo{person}{Faegheh Hasibi}, \bibinfo{person}{Jiaxin Mao}, {and} \bibinfo{person}{Joemon~M. Jose}} (Eds.). \bibinfo{publisher}{{ACM}}, \bibinfo{pages}{259--268}.
\newblock
\urldef\tempurl%
\url{https://doi.org/10.1145/3673791.3698414}
\showDOI{\tempurl}


\bibitem[Zhuang et~al\mbox{.}(2024a)]%
        {zhuang2024doesvec2textposenew}
\bibfield{author}{\bibinfo{person}{Shengyao Zhuang}, \bibinfo{person}{Bevan Koopman}, {and} \bibinfo{person}{Guido Zuccon}.} \bibinfo{year}{2024}\natexlab{a}.
\newblock \showarticletitle{Does Vec2Text Pose a New Corpus Poisoning Threat?}
\newblock \bibinfo{journal}{\emph{CoRR}}  \bibinfo{volume}{abs/2410.06628} (\bibinfo{year}{2024}).
\newblock
\urldef\tempurl%
\url{https://doi.org/10.48550/ARXIV.2410.06628}
\showDOI{\tempurl}


\end{thebibliography}

\end{document}